\documentclass{article}
\usepackage[final]{iclr2025_conference}  
\usepackage{times}

\usepackage{hyperref}
\usepackage{url}
\usepackage{booktabs}
\usepackage{graphicx}
\usepackage{amsmath,amssymb}
\usepackage{math_commands}
\usepackage{xcolor}
\usepackage{algorithm}
\usepackage{algorithmic}
\usepackage{subcaption}
\usepackage{multirow}
\usepackage{float}
\usepackage{tikz}
\usetikzlibrary{arrows.meta}

\newcommand{\sign}{\operatorname{sign}}

\newcommand{\method}{SymExpLin\xspace}
\newcommand{\methodshort}{SEL\xspace}

\title{Learning in Curved Weight Space:\\Exponential-Linear Weight Reparameterization for Improved Optimization}

\author{%
Ethan Smith \\
Canva Research \\
\texttt{ethansmith@canva.com}
}

\begin{document}

\maketitle

\begin{abstract}
Many neural networks operations have a multiplicative nature rather than additive: halving or doubling a norm are analogous relatively but require unequal optimization distances when taking linear steps.
Adaptive optimizers such as Adam normalize updates per coordinate, but update steps remain additive; weights with very different magnitudes receive similarly sized absolute changes, producing very different relative perturbations.
We introduce \textbf{\method} (\textbf{\methodshort}), a weight reparameterization for neural networks that combines a sign-aware symmetric-exponential pathway with an identity-like linear pathway.
The symmetric-exponential pathway is near-linear for small raw weights but increasingly curved at larger magnitudes. Additive updates in logarithmic space map to magnitude-proportional changes in effective weight space.
The linear pathway provides a direct route through the transform, while learnable scale, curvature, and offset parameters control pathway balance and exponential curvature.
Controlled small-model ablations show that the curved parameterization improves optimization even with these auxiliary scales frozen, that learned pathway scales can provide a further gain, and that an optimizer-side implementation of frozen exponential-only SEL closely reproduces its trajectory.
We also identify a useful \emph{mismatched initialization}: raw weights are chosen so a symmetric version of the transform matches Xavier statistics, but training uses an asymmetric forward transform that leaves positive weights at full strength while making negative weights smaller in magnitude; in small-model ablations, this improves early optimization and may act as a form of symmetry breaking.
We train transformers on OpenWebText over nine width$\times$depth configurations, \methodshort reaches matched validation loss in 1.32--1.49$\times$ fewer training steps, with the largest widths seeing the biggest gains.
\methodshort is practical: in our largest profiled configuration, a 1.44$\times$ step reduction and 5.5\% per-step overhead correspond to an estimated 1.37$\times$ wall-clock speedup, and after training the parameterization is folded into standard linear weights with no inference cost.
\end{abstract}

\section{Introduction}

Standard neural network training optimizes parameters in an additive fashion with linear update steps.
Yet many computations within a network are naturally relative. Doubling or halving a normalization gain, a feature amplitude, or the singular value of a matrix reflects analogous changes in logarithmic space. However, they have unequal optimization distances in linear weight space. More specifically, moving from $1$ to $0.5$ has half the linear distance of moving from $1$ to $2$. 
This is not to say we should necessarily think of all aspects of the network in a multiplicative fashion. Activation functions alongside biases consider absolute scale to be important. Nevertheless, the optimization employed in training has mostly played into the linear/absolute scale nature of networks. When relative changes weights are to be achieved through additive updates, we should expect training to push a subset of parameters far from initialization whenever the task demands large multiplicative factors, while leaving many others near their starting scale.

As shown in Figure~\ref{fig:weight_dist}, we may observe this effect in pretrained language models, which develop heavy-tailed weight distributions. Many parameters remain near initialization scale, while a small fraction move far into the tails.
In our Qwen3-4B-Instruct analysis (Figure~\ref{fig:weight_dist}), the maximum weight is roughly 30$\times$ the Xavier initialization scale even though the standard deviation grows only 1.7$\times$.
This creates a heterogeneous optimization problem in which many coordinates may travel little from their original positions, while some coordinates must traverse much larger distances in weight space.

Adaptive optimizers such as Adam~\citep{kingma2014adam} normalize update magnitudes per coordinate inducing similar magnitude across dimensions, rendering the updates invariant to parameter magnitude.
Preconditioning decouples step size from gradient scale to keep optimization well-conditioned, but a parameter at $0.004$ and one at $0.4$ can receive similar absolute updates while resulting in very different relative perturbations: the same step may be significant for a small parameter yet barely move a large one in relative terms.
Global learning rates may be chosen to respect the most sensitive small-scale coordinates, potentially underserving larger magnitude directions such as feature gains, matrix modes, or other coordinates that may benefit from relative movement.

One solution may be to try to correct this at the optimizer level through parameter-magnitude scaling into the update rule, but this may fight against the preconditioning benefit. We then propose to change the parameterization (and thus geometry) of the weights themselves. The natural method for incorporating multiplicative relativity is to make use of an exponential mapping, which converts additive additive updates in the raw space into multiplicative updates in the output space. However, the standard exponential function only outputs positive values when our weights are roughly centered and symmetrical around 0. To address this issue, we utilize the symexp function, which is sign-preserving, odd, monotonic, and offers the exponential behavior.
As to not overcommit to the exponential structure, we introduce a linear pathway along with learnable weights allowing the network to scale the effect of expoential and linear components. We hypothesize as well that the linear pathway may provide a direct gradient route which may aid optimization.

\textbf{Contributions:}
\begin{itemize}
    \item We propose \method (\methodshort), a nonlinear weight reparameterization applicable to any learnable weight matrix or bias vector, decomposing effective weights into symmetric-exponential and linear components with independently learnable structured scales.
    \item We identify \emph{mismatched initialization}, using a different combination rule for weight inversion than for the forward pass, as a source of beneficial asymmetric bias in our small-scale ablations.
    \item We introduce structured scale parameters with various factorization patterns and a targeted learning rate annealing schedule for managing redundant exponential controls.
    \item Controlled mechanism ablations separate fixed curved geometry from learned scales and structured scaling alone; an optimizer-coordinate control reproduces frozen exponential-only SEL without changing the forward parameterization.
    \item \methodshort is implementation-friendly: the training transform is elementwise and incurs modest overhead in compiled implementations, while effective weights are precomputed after training, reducing to standard linear layers with \textbf{zero inference cost}.
    \item We note that \methodshort can be folded into standard weights after training and, conversely, standard weights can be inverted back into the \methodshort parameter space for future continued-training or finetuning experiments.
    \item We evaluate across a grid of model scales ($\{1024, 2048, 3072\} \times \{12, 24, 36\}$) on autoregressive language modeling, with targeted ablations for initialization asymmetry, the linear pathway, and scale learning-rate annealing.
\end{itemize}

\begin{figure*}[t]
\centering
\begin{tikzpicture}[
    x=1cm,
    y=1cm,
    font=\sffamily,
    panel/.style={
        draw=black!22,
        fill=black!2,
        rounded corners=3pt,
        minimum width=5.25cm,
        minimum height=4.35cm,
    },
    header/.style={font=\sffamily\bfseries\small, text=black!78, align=center},
    body/.style={font=\sffamily\scriptsize, text=black!72, align=center},
    mini/.style={
        draw=black!18,
        rounded corners=2pt,
        minimum width=1.42cm,
        minimum height=0.72cm,
        font=\sffamily\scriptsize,
        align=center,
    },
    flow/.style={-{Stealth[length=2mm]}, line width=0.7pt, black!52},
    curve/.style={line width=1.4pt, violet!78!black},
]
    \node[font=\sffamily\bfseries\large, text=black!82] at (8.25,5.15)
        {Train in curved weight space. Deploy standard linear layers.};

    \node[panel] at (2.75,2.55) {};
    \node[panel] at (8.25,2.55) {};
    \node[panel] at (13.75,2.55) {};

    \node[header] at (2.75,4.22) {1. Map weights into SEL form};
    \begin{scope}[shift={(0.82,2.42)}]
        \draw[fill=blue!7, draw=blue!34, rounded corners=1pt] (0,0) rectangle (1.32,1.32);
        \foreach \x/\y/\shade in {
            0.06/0.06/42, 0.48/0.06/17, 0.90/0.06/30,
            0.06/0.48/20, 0.48/0.48/48, 0.90/0.48/12,
            0.06/0.90/34, 0.48/0.90/14, 0.90/0.90/40
        } {
            \draw[fill=blue!\shade, draw=white] (\x,\y) rectangle ++(0.36,0.36);
        }
    \end{scope}
    \node[body, font=\sffamily\scriptsize\bfseries, text=blue!48!black]
        at (1.48,2.12) {Linear weights};

    \draw[flow] (2.30,3.08) -- (3.20,3.08);
    \node[font=\sffamily\tiny, text=black!58] at (2.75,3.39)
        {invert $T_\theta$};

    \begin{scope}[shift={(3.36,2.42)}]
        \draw[fill=violet!7, draw=violet!36, rounded corners=1pt] (0,0) rectangle (1.32,1.32);
        \foreach \x/\y/\shade in {
            0.06/0.06/42, 0.48/0.06/17, 0.90/0.06/30,
            0.06/0.48/20, 0.48/0.48/48, 0.90/0.48/12,
            0.06/0.90/34, 0.48/0.90/14, 0.90/0.90/40
        } {
            \draw[fill=violet!\shade, draw=white] (\x,\y) rectangle ++(0.36,0.36);
        }
    \end{scope}
    \node[body, font=\sffamily\scriptsize\bfseries, text=violet!62!black]
        at (4.02,2.12) {SEL parameters};

    \node[body, text=black!62] at (2.75,1.02)
        {The inverse transform preserves effective\\weight values in the SEL representation};

    \node[header] at (8.25,4.22) {2. Train in curved weight space};
    \begin{scope}[shift={(8.25,2.70)}]
        \draw[black!34, line width=0.45pt, -{Stealth[length=1.3mm]}] (-1.72,0) -- (1.78,0);
        \draw[black!34, line width=0.45pt, -{Stealth[length=1.3mm]}] (0,-0.86) -- (0,1.12);
        \draw[densely dashed, black!32, line width=0.55pt] (-1.52,-0.58) -- (1.52,0.58);
        \draw[curve] (-1.52,-0.96)
            .. controls (-1.18,-0.55) and (-0.50,-0.13) .. (0,0)
            .. controls (0.50,0.13) and (1.18,0.55) .. (1.52,0.96);

        \node[font=\sffamily\tiny, text=black!52] at (1.62,-0.17) {$w$};
        \node[font=\sffamily\tiny, text=violet!72!black] at (-0.52,1.02)
            {$w_{\mathrm{eff}}$};

        \draw[blue!55!black, {Stealth[length=1.1mm]}-{Stealth[length=1.1mm]}, line width=0.65pt]
            (0.16,-0.55) -- (0.46,-0.55);
        \draw[blue!55!black, {Stealth[length=1.1mm]}-{Stealth[length=1.1mm]}, line width=0.65pt]
            (0.88,-0.55) -- (1.18,-0.55);
        \node[font=\sffamily\tiny, text=blue!48!black] at (0.31,-0.74) {$\Delta w$};
        \node[font=\sffamily\tiny, text=blue!48!black] at (1.03,-0.74) {$\Delta w$};

        \fill[orange!82!black] (0.16,0.04) circle (1.4pt);
        \fill[orange!82!black] (0.46,0.13) circle (1.4pt);
        \fill[orange!82!black] (0.88,0.34) circle (1.4pt);
        \fill[orange!82!black] (1.18,0.62) circle (1.4pt);
        \draw[orange!82!black, -{Stealth[length=1.5mm]}, line width=1.5pt]
            (0.16,0.04) -- (0.46,0.13);
        \draw[orange!82!black, -{Stealth[length=1.5mm]}, line width=1.5pt]
            (0.88,0.34) -- (1.18,0.62);

    \end{scope}
    \node[body, text=black!62, text width=4.45cm] at (8.25,0.90)
        {Equal additive steps $\Delta w$ yield larger effective-weight movements as $\abs{w}$ grows};

    \node[header] at (13.75,4.22) {3. Materialize linear weights\\for inference};
    \begin{scope}[shift={(11.70,2.42)}]
        \draw[fill=violet!7, draw=violet!36, rounded corners=1pt] (0,0) rectangle (1.32,1.32);
        \foreach \x/\y/\shade in {
            0.06/0.06/42, 0.48/0.06/17, 0.90/0.06/30,
            0.06/0.48/20, 0.48/0.48/48, 0.90/0.48/12,
            0.06/0.90/34, 0.48/0.90/14, 0.90/0.90/40
        } {
            \draw[fill=violet!\shade, draw=white] (\x,\y) rectangle ++(0.36,0.36);
        }
    \end{scope}
    \node[font=\sffamily\scriptsize\bfseries, text=violet!62!black, align=center]
        at (12.36,2.12) {SEL parameters};

    \draw[flow] (13.18,3.08) -- (14.30,3.08);
    \node[font=\sffamily\tiny, text=black!58] at (13.74,3.39)
        {$T_\theta$};

    \begin{scope}[shift={(14.46,2.42)}]
        \draw[fill=blue!7, draw=blue!34, rounded corners=1pt] (0,0) rectangle (1.32,1.32);
        \foreach \x/\y/\shade in {
            0.06/0.06/42, 0.48/0.06/17, 0.90/0.06/30,
            0.06/0.48/20, 0.48/0.48/48, 0.90/0.48/12,
            0.06/0.90/34, 0.48/0.90/14, 0.90/0.90/40
        } {
            \draw[fill=blue!\shade, draw=white] (\x,\y) rectangle ++(0.36,0.36);
        }
    \end{scope}
    \node[font=\sffamily\scriptsize\bfseries, text=blue!48!black, align=center]
        at (15.12,2.12) {Linear weights};

    \node[
        draw=green!45!black,
        fill=green!10,
        rounded corners=2pt,
        minimum width=3.85cm,
        minimum height=0.62cm,
        font=\sffamily\scriptsize\bfseries,
        text=green!32!black,
        align=center,
    ] at (13.75,1.02) {zero inference overhead};

    \node[
        draw=blue!50!black,
        fill=blue!7,
        rounded corners=3pt,
        minimum width=10.3cm,
        minimum height=0.6cm,
        font=\sffamily\small\bfseries,
        text=blue!48!black,
        align=center,
    ] at (8.25,-0.22)
        {1.32--1.49$\times$ fewer training steps to matched validation loss observed in experiments};
\end{tikzpicture}
\caption{\textbf{SymExpLin in one view.} Existing linear weights are mapped into SEL form without changing their effective values. Training then proceeds in the curved parameterization, where additive raw-weight updates induce magnitude-dependent movement in effective weight space. After training, the transform is evaluated once to materialize ordinary linear weights with no additional inference cost.}
\label{fig:sel_overview}
\end{figure*}

\section{Related Work}
\label{sec:related}

\paragraph{Weight Reparameterization.}
Weight normalization~\citep{salimans2016weight} separates weight magnitude from direction, improving optimization conditioning.
Spectral normalization~\citep{miyato2018spectral} constrains the spectral norm for training stability.
Both apply relatively simple transformations.
\methodshort introduces a richer nonlinear reparameterization that mixes symmetric-exponential and linear components with learnable mixture coefficients, operating on the weight matrix itself rather than its norms.

\paragraph{Symlog Transforms.}
The symmetric logarithm $\sign(x)\ln(\abs{x}+1)$ and its inverse have been used for prediction targets in reinforcement learning~\citep{hafner2023mastering}, handling values spanning many orders of magnitude.
\methodshort applies a similar operation, a parameterized symmetric exponential, to \emph{weight space} rather than prediction targets, with learnable scale parameters controlling the symmetric-exponential and linear components.

\paragraph{Initialization Schemes.}
Xavier~\citep{glorot2010understanding} and Kaiming~\citep{he2015delving} initialization maintain variance across layers.
Data-dependent initialization~\citep{mishkin2016all} matches target statistics layer by layer.
Our initialization via Newton's method inversion ensures that \emph{despite} the nonlinear reparameterization, effective weights at initialization match standard Xavier statistics.
Our mismatched initialization is different from these schemes: it preserves the intended Xavier scale under one transform rule, then deliberately evaluates the same raw weights under a slightly asymmetric rule.

\paragraph{Learning Rate Parameterization.}
Maximal Update Parameterization ($\mu$P)~\citep{yang2022tensor} provides principled per-tensor learning rate scaling across model widths, addressing one axis of the init/optimizer mismatch we discuss in Section~\ref{sec:motivation}. \methodshort targets a complementary axis, adapting effective step size per-parameter over the course of training rather than per-tensor at initialization; the two are not mutually exclusive.

\paragraph{Multiplicative Interactions.}
The Multiplicative Weights Update (MWU) method~\citep{arora2012multiplicative} makes relative-scale updates explicit: weights are scaled by $(1 \pm \epsilon)^{\text{payoff}}$ rather than shifted additively, so adjustments are naturally proportional to current magnitude.
This principle underlies algorithms from boosting to online convex optimization, but has seen limited adoption in neural network training, where additive SGD-style updates remain standard.
\methodshort does not update multiplicatively.
The optimizer still takes additive steps in raw parameter space, but the symmetric-exponential pathway induces a related effect at the effective-weight level: raw-space sensitivity scales as $\exp(\beta\abs{w})$, so identical raw steps produce larger effective-weight displacements at larger magnitudes.
The linear pathway preserves additive behavior near the origin, yielding a hybrid regime rather than a fully multiplicative one.
This positions \methodshort as a way to obtain some of MWU's relative-scale behavior while remaining compatible with standard gradient-based optimizers and preconditioning.

\section{Motivation}
\label{sec:motivation}

\begin{figure}[t]
\centering
\includegraphics[width=0.95\textwidth]{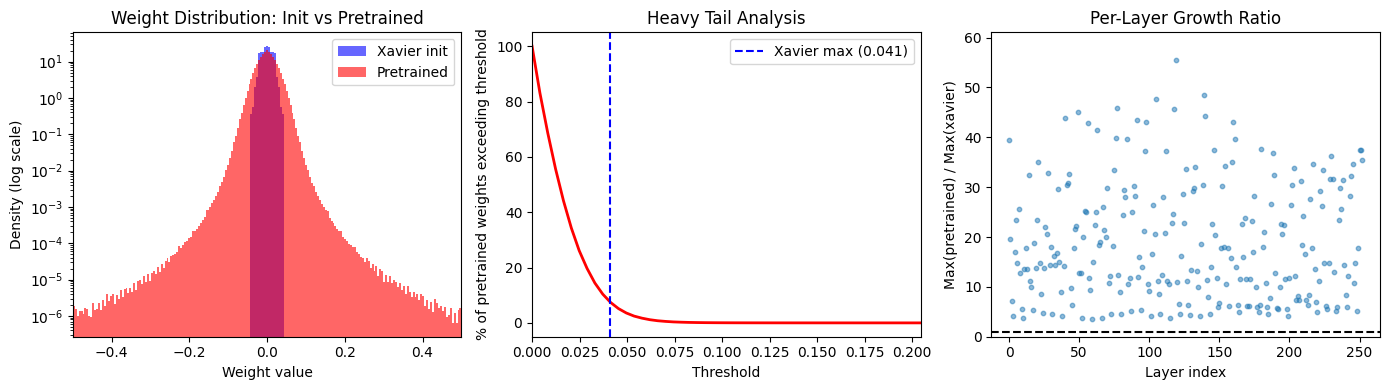}
\caption{\textbf{Weight distribution gap between initialization and pretrained models.} \emph{Left:} Histogram comparison (log scale) showing pretrained weights have heavier tails than Xavier initialization. \emph{Middle:} Tail analysis. The fraction of pretrained weights exceeding various thresholds. The vertical line marks the Xavier max. \emph{Right:} Per-layer growth ratio; most layers have weights that grew 5--15$\times$ beyond their initialization range.}
\label{fig:weight_dist}
\end{figure}

Training moves parameters from a sampled simple starting distribution to a much broader final one. 
At initialization, variance-preserving schemes initalize parameters near a typical magnitude determined by its matrix's width. 
By the end of training, we observe that the distribution of weights moves towards a heavy tailed shape in which a small fraction of parameters live far outside the initial typical range (Figure~\ref{fig:weight_dist}). 
Different families of methods respond to different points on this trajectory: $\mu$P and related learning-rate parameterizations set step size from the init-side typical magnitude, a per-tensor summary statistic downstream of model dimension, while \methodshort adapts effective step size dynamically from each parameter's \emph{current} magnitude, so as the distribution spreads during training, parameters that have travelled further receive proportionally larger effective steps. 
This section discusses preconditioning, learning rate tuning, and update magnitude with respect to parameter magnitude. We pose \methodshort as a complementary counterpart to $\mu$P along a second, dynamic axis.

\paragraph{Adam normalizes away scale.}
Consider Adam's update rule.
The first and second moment estimates are $m_t = \beta_1 m_{t-1} + (1-\beta_1) g_t$ and $v_t = \beta_2 v_{t-1} + (1-\beta_2) g_t^2$, and the update is $\Delta w \propto m_t / \sqrt{v_t}$.
Under steady-state conditions with approximately constant gradient magnitudes, $v_t \approx \E[g^2]$, so the normalized update $m_t / \sqrt{v_t}$ has expected absolute value near 1 given the division by the estimated gradient's standard deviation.
In the special case $\beta_1 = \beta_2$ where the two moving averages track on identical timescales, Cauchy-Schwarz tightens this to a hard bound: $\abs{m_t / \sqrt{v_t}} \leq 1$ at steady state.
Beyond steady state conditions, momentum suffers destructive interference under gradient sign flips, decreasing the numerator. Meanwhile $v_t$ is sensitive to larger values potentially increasing denominator.
The essential point is invariance: uniformly scaling every gradient by a constant leaves $\abs{m_t / \sqrt{v_t}}$ unchanged, the normalization by $\sqrt{v_t}$ cancels it, and no term in the update depends on $\abs{w}$. Per-dimension variation reflects the \emph{shape} of the gradient distribution (noise pattern, outlier frequency) rather than its overall scale, so updates concentrate around a common magnitude set by the learning rate, invariant to both expected gradient magnitude and parameter magnitude.
A parameter at $0.004$ and one at $0.4$ therefore receive updates of similar absolute size, corresponding to a $100\times$ difference in relative perturbation; global learning rate tuning must accommodate the most sensitive small-scale coordinates, potentially under-serving larger-magnitude ones.

\paragraph{Variance-Preserving Initialization}
Variance-preserving initialization offer a typical parameter magnitude at the start of training, and may also guide common magnitudes at the end state of training.
Xavier initialization uses $\sigma \propto 1/\sqrt{(\text{fan\_in} + \text{fan\_out})/2}$, so smaller matrices receive larger initialization values and larger matrices smaller ones.
The effect is pronounced for low-rank adapters~\citep{hu2021lora}, where the rank $r \ll d$ yields much larger initialization scales than full weight matrices. This is consistent with the observation that LoRA requires higher learning rates than full model weight finetuning.
Initialization thus writes scale information into the parameters that Adam then normalizes away.

\paragraph{$\mu$P: per-tensor, width-driven, static.}
Maximal Update Parameterization~\citep{yang2022tensor} formalizes one axis of this mismatch: typical init magnitude varies systematically with matrix width, and per-tensor learning rates must be rescaled with width to preserve feature-learning behavior across scales.
$\mu$P's rescaling is theoretically derived, applied once at the tensor level, and driven by static properties of the architecture. In effect, it utilizes either the standard deviation or uniform bound implied by the width and adjusts step size accordingly.

\paragraph{A second, dynamic axis.}
$\mu$P addresses variation \emph{across} tensors of different widths but not variation \emph{within} a tensor over training.
Xavier init specifies a typical starting scale for each parameter, but pretrained networks develop heavy-tailed weight distributions in which a small fraction of parameters end up an order of magnitude beyond that typical scale.
We verified this on Qwen3-4B-Instruct~\citep{qwen3}: while Xavier initialization produces weights with $\max \approx 0.04$ and $\sigma \approx 0.014$, the pretrained weights have $\max \approx 1.23$ (a $30\times$ increase) while $\sigma$ increases only $1.7\times$.
The distribution is \emph{heavy-tailed}: only $0.05\%$ of weights exceed $0.1$ in magnitude, yet these outliers are functionally critical~\citep{dettmers2022llm} (Figure~\ref{fig:weight_dist}).
Under Adam updates, or other optimizer with normalization, some parameters fundamentally have greater distance to travel requiring more optimization steps.

\paragraph{\methodshort as a per-parameter, adaptive analog.}
\methodshort takes a similar underlying observation as $\mu$P and pushes it along this second axis.
Instead of adjusting learning rates at a tensor or model-level from static init statistics, we change the parameterization sucht that the local sensitivity $\partial w_{\mathrm{eff}} / \partial w_{\mathrm{raw}}$ grows with $\abs{w}$. Identical raw-space steps produce larger effective-weight movements for larger parameters.
In this sense \methodshort is a per-parameter, dynamic, empirically motivated counterpart to $\mu$P's per-tensor, static, theoretically motivated rescaling. $\mu$P derives scale from typical parameter value obtained from width, \methodshort obtain scale from individual dimensions throughout training.
The two operate on distinct axes, are not mutually exclusive, and can be composed where both apply.

\section{Method}
\label{sec:method}

\subsection{Background: SymLog and SymExp}

The symmetric logarithm (symlog) and its inverse, the symmetric exponential (symexp), are odd functions that handle signed values spanning many orders of magnitude~\citep{hafner2023mastering}:
\begin{align}
\operatorname{symlog}(x) &= \sign(x) \ln(\abs{x} + 1) \\
\operatorname{symexp}(x) &= \sign(x) \bigl(\exp(\abs{x}) - 1\bigr)
\end{align}
The symexp function is particularly relevant. For small $\abs{x}$, it behaves approximately linearly ($\operatorname{symexp}(x) \approx x$), while for large $\abs{x}$, it grows exponentially.
It is sign-preserving and continuously differentiable, but not twice differentiable at the origin (Figure~\ref{fig:symexp_controls}, left).

A natural idea is to use symexp as a weight reparameterization. Store raw parameters $w$ and use $\operatorname{symexp}(w)$ as the effective weight.
This creates a curved optimization landscape where large weights are amplified.
However, bare symexp has a fixed curvature and no mechanism to independently control the balance between its near-linear small-weight regime and its exponential large-weight regime.

\method generalizes this idea by (1) introducing a curvature parameter $\beta$ that controls the transition between linear and exponential regimes, (2) separating the exponential and linear contributions into independently scalable pathways, and (3) adding learnable modulation parameters that control curvature, thresholding, and pathway balance.

\subsection{The \method Transform}

Given a raw weight parameter $w \in \R$, \method generalizes symexp by introducing a curvature parameter $\beta$ and a linear pathway that we hypothesize may aid optimization.
We first give a canonical \emph{congruent} form, before introducing the mismatch combination used in our main configuration:
\begin{equation}
\label{eq:base_transform}
w_{\mathrm{eff}} = \frac{\sign(w) \cdot \bigl(\exp(\beta\abs{w}) - 1\bigr)}{\beta} + \frac{w}{\beta}
\end{equation}
The first term is a scaled symexp. Placing $\beta$ inside the exponent sets the exponential growth rate, while the outer $1/\beta$ normalizes the leading-order Taylor expansion so the symexp itself contributes unit slope at the origin regardless of $\beta$ (since $\exp(\beta \abs{w}) - 1 \approx \beta \abs{w}$ near zero).
The linear pathway is likewise divided by $\beta$, so its contribution is proportional to $1/\beta$: it adds a small $1/\beta$ slope near the origin that fades as $\beta$ grows and the exponential regime dominates.
Thus, the total near-zero slope of the congruent base form is $1+1/\beta$, not exactly one.
The normalization keeps this deviation small at the effective curvature values used in our experiments: it is at most $1/7.5 \approx 0.133$ when $\beta \geq 7.5$, and vanishes as $\beta$ grows.
Consequently, changing $\beta$ primarily changes curvature and the relative balance of the two pathways, rather than introducing an unconstrained change in overall gain.

\begin{figure}[t]
\centering
\includegraphics[width=\textwidth]{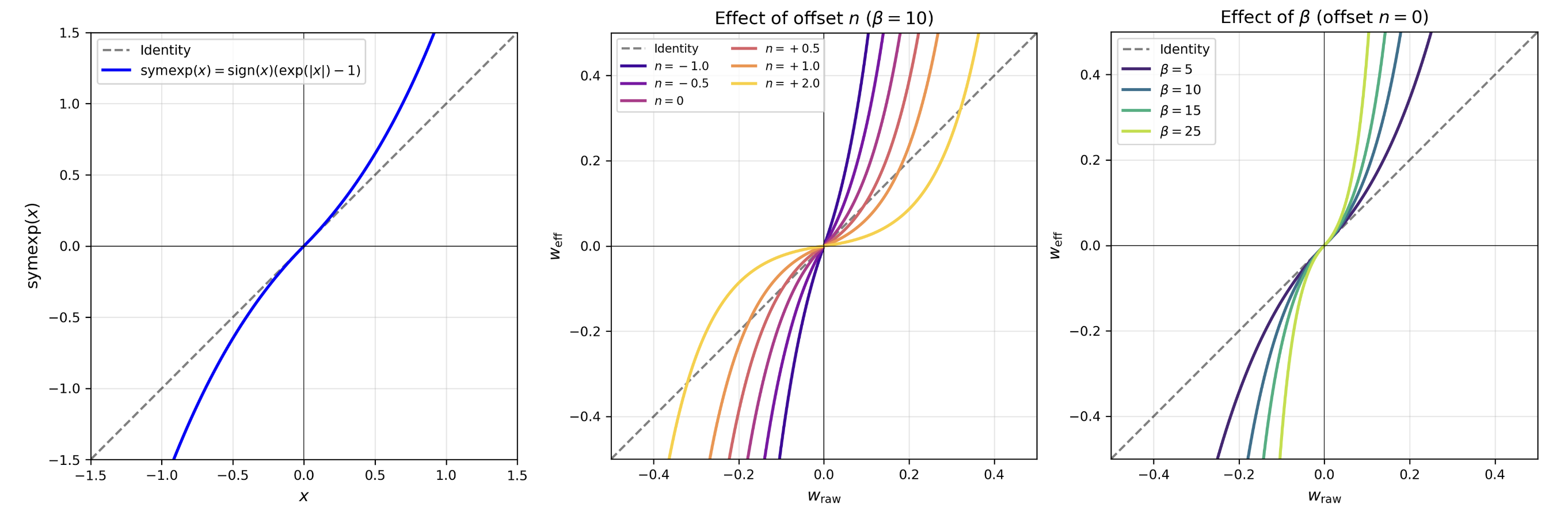}
\caption{\textbf{Symmetric exponential transform and controls.} \emph{Left:} The standard symexp function is approximately linear near zero and diverges exponentially from the identity at larger magnitudes while preserving sign. \emph{Middle:} Under the default normalized exponential pathway, the offset $n$ sets the slope at the origin via $\exp(-n)$: $n = 0$ gives unit slope, positive $n$ shrinks the slope so the transform starts flatter and reaches the exponential regime later, and negative $n$ raises the initial slope so amplification begins earlier. \emph{Right:} The curvature $\beta$ is the exponential growth rate: larger $\beta$ produces a sharper transition into the exponential regime and faster growth within it. Since Xavier initialization magnitudes shrink with model width, larger models generally need larger $\beta$ to reach meaningful curvature at typical weight magnitudes.}
\label{fig:symexp_controls}
\end{figure}

The choice of $\beta$ should be calibrated to the typical initialization scale.
Recall that Xavier initialization uses $\sigma \propto 1/\sqrt{d}$, so larger matrices have smaller initial weight magnitudes.
To achieve meaningful curvature within this range, $\beta$ must be correspondingly larger, otherwise the transform remains extensively in its near-linear regime.
Figure~\ref{fig:symexp_controls} (right) illustrates this relationship.

\paragraph{Adding learnable control.}
\method separates the coefficient that controls exponential curvature from the shared pathway normalizer.
Let $\kappa$ denote the effective curvature and $d$ the normalizer:
\begin{equation}
\label{eq:curvature_normalizer}
(\kappa, d) =
\begin{cases}
(\beta m, \beta), & \text{fixed-$\beta$, unnormalized}, \\
(\beta m, \beta m), & \text{fixed-$\beta$, normalized}, \\
(\tilde{\beta}, \beta), & \text{absorbed-$\beta$, unnormalized}, \\
(\tilde{\beta}, \tilde{\beta}), & \text{absorbed-$\beta$, normalized}.
\end{cases}
\end{equation}
Here $m$ is a learnable inner multiplier initialized near one, while $\tilde{\beta}$ is a directly learned, positive effective curvature initialized at $\beta$.
The base transform in Eq.~\eqref{eq:base_transform} is the special case $\kappa=d=\beta$ with all learnable scales set to one.
For readability, we write scalar equations; in implementation, these quantities may be structured scales over a weight matrix.

The \textbf{symmetric-exponential pathway} and the \textbf{linear pathway} are:
\begin{equation}
\label{eq:exp_path}
\begin{aligned}
E_\theta(\abs{w}) &= \frac{e_w}{d}\Bigl(\exp\bigl(\kappa\abs{w} - n\bigr) - \exp(-n)\Bigr), \\
S_\theta(w) &= \sign(w) \cdot E_\theta(\abs{w}), \\
L_\theta(w) &= \frac{l_w}{d}w.
\end{aligned}
\end{equation}
Here $E_\theta(\abs{w})$ is the exponential amplitude, $S_\theta(w)$ is the signed symmetric-exponential pathway, $n$ is a learnable offset, and $e_w,l_w$ are learnable relative pathway scales.
Writing $1/d$ explicitly matches the implementation: $e_w$ and $l_w$ initialize to one, which keeps their initialization and optimization geometry consistent as the selected base curvature changes across network sizes.

The offset $n$ sets the exponential path's local slope:
\begin{equation}
\left.\frac{\partial E_\theta}{\partial \abs{w}}\right|_{0}
= e_w \frac{\kappa}{d}\exp(-n).
\end{equation}
Therefore, effective-curvature normalization ($d=\kappa$) preserves this slope as $e_w\exp(-n)$ while $\kappa$ changes the curvature and growth rate.
In the unnormalized fixed-$\beta$ form, the corresponding slope is $e_w m\exp(-n)$.
The default $n = 0$ leaves the exponential path's local slope at $e_w\kappa/d$; positive $n$ shrinks this slope, making the transform flatter near zero and delaying when the exponential regime takes over, while negative $n$ raises it so amplification begins earlier.
The two appearances of $n$ in Equation~\eqref{eq:exp_path}, as $-n$ inside the exponent and as $-\exp(-n)$ outside, are what preserve $E_\theta(0) = 0$ while allowing $n$ to modulate this slope; the effect is the single-parameter reshaping just described.
Figure~\ref{fig:symexp_controls} (middle) visualizes this effect.

Although $d=\kappa$ fixes the exponential path's local gain, it changes the linear contribution to $l_w/\kappa$.
With $e_w=l_w=1$ and $n=0$, the congruent local slope is $1+1/\kappa$; in mismatch mode, the positive and negative one-sided slopes are $1+1/\kappa$ and $1-1/\kappa$, respectively.
For $\kappa \geq 7.5$, this residual linear contribution is at most $13.3\%$ of the unit exponential gain and tends to zero as $\kappa$ grows.

A learnable residual $\Delta_\phi$ optionally provides an additive correction.
In our experiments, this residual is typically implemented as a low-rank LoRA-style update and can be omitted without changing the core \methodshort parameterization.
The \textbf{full transform} in \emph{mismatch mode} (our recommended configuration) is:
\begin{equation}
\label{eq:full_transform}
w_{\mathrm{eff}} = \sign(w) \cdot \bigl(E_\theta(\abs{w}) + L_\theta(w)\bigr) + \Delta_\phi
\end{equation}

\paragraph{Role of each component.}
The reference curvature $\beta$ is chosen to match the initialization scale.
The effective curvature $\kappa$ controls the transition between linear and exponential regimes, either through the fixed-$\beta$ multiplier $m$ or through the directly learned $\tilde{\beta}$.
The normalizer $d$ determines whether changes in effective curvature are also normalized in both pathways.
The offset $n$ controls where exponential amplification begins.
The scales $e_w$ and $l_w$ independently weight each pathway's contribution.
When enabled, the residual $\Delta_\phi$ adds a sign-independent correction; its learning rate is scaled by $1/\beta$ to keep its effective step size commensurate with the other terms.

\begin{figure}[t]
\centering
\includegraphics[width=0.95\textwidth]{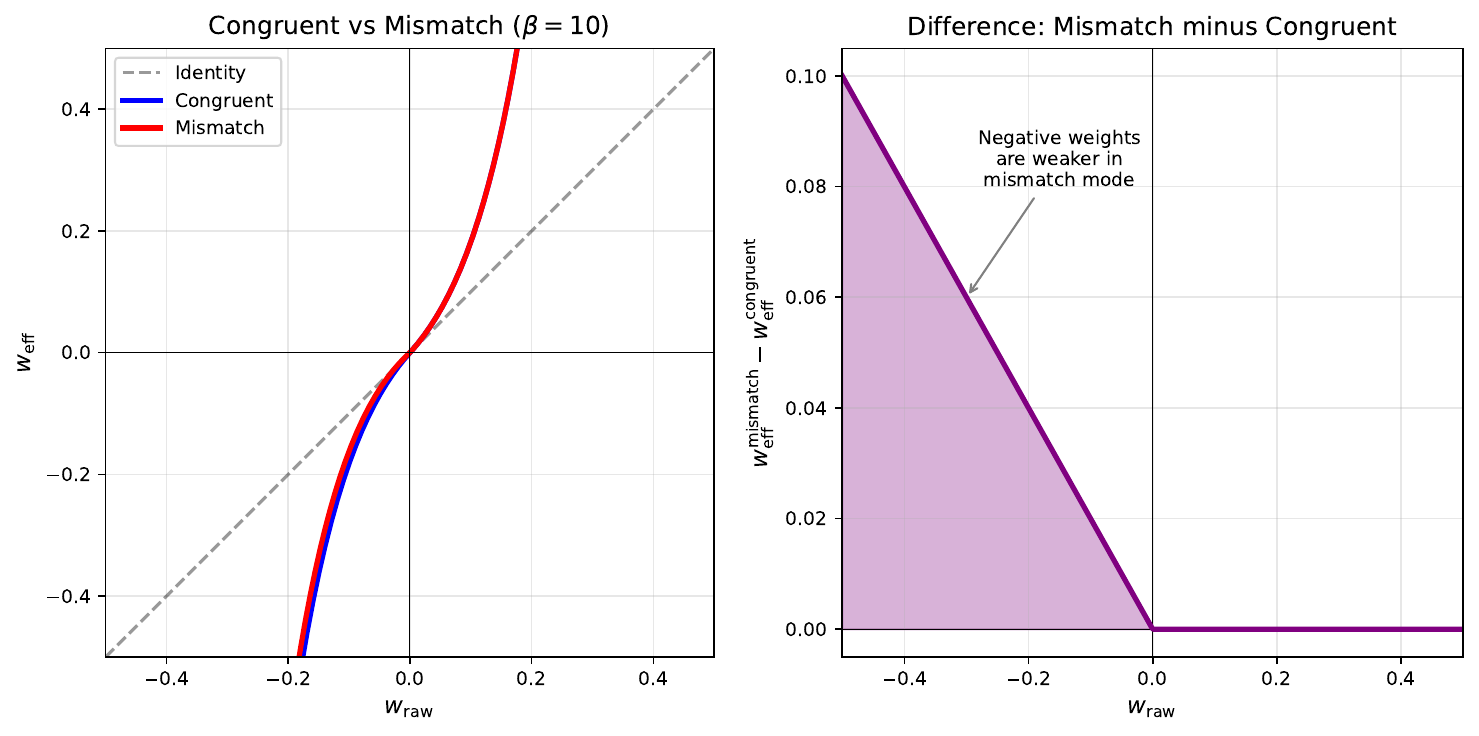}
\caption{\textbf{Congruent vs mismatch combination modes.} \emph{Left:} Both modes overlaid. For positive $w$, the curves coincide. For negative $w$, mismatch mode (red) produces smaller effective magnitudes than congruent mode (blue). \emph{Right:} The difference between modes. The asymmetry arises because in mismatch mode, the linear term $L_\theta(w)$ (which is negative when $w < 0$) is wrapped inside the sign operation, reducing the magnitude rather than reinforcing it.}
\label{fig:congruent_vs_mismatch}
\end{figure}

\paragraph{Comparison of combination modes.}
We define two modes for combining the pathways:
\begin{align}
\text{Congruent:} \quad w_{\mathrm{eff}} &= S_\theta(w) + L_\theta(w) \label{eq:congruent} \\
\text{Mismatch:} \quad w_{\mathrm{eff}} &= \sign(w) \cdot \bigl(E_\theta(\abs{w}) + L_\theta(w)\bigr) \label{eq:mismatch} \\
\end{align}
In congruent mode~\eqref{eq:congruent}, the linear part is added \emph{outside} the sign wrapping and thus independently preserves $w$'s sign.
In mismatch mode~\eqref{eq:mismatch}, both parts are inside the sign wrapping: for negative $w$, the linear term (which is negative) reduces the magnitude.
Figure~\ref{fig:congruent_vs_mismatch} illustrates this difference: the two modes produce identical outputs for positive weights, but for negative weights, mismatch mode produces \emph{smaller} effective magnitudes than congruent mode.
This asymmetry is one mechanism we study for improving early optimization.
In our width-1024 ablations, mismatch mode outperforms congruent, particularly when combined with congruent initialization (\S\ref{sec:mismatch}).

\subsection{Mismatched Initialization}
\label{sec:mismatch}

To ensure that effective weights at initialization match standard Xavier statistics~\citep{glorot2010understanding}, we initialize the raw weights by \emph{inverting} the transform: given target weights $W_{\mathrm{target}}$ from Xavier uniform initialization, we solve $f(w) = W_{\mathrm{target}}$ via Newton's method.

The key design choice is \emph{which} combination rule to use for this inversion.
In our width-1024 ablations, using the \textbf{congruent formula}~\eqref{eq:congruent} for inversion while using the \textbf{mismatch formula}~\eqref{eq:mismatch} for the forward pass yields the strongest results.
We call this \emph{mismatched initialization}.

Under congruent inversion, Newton's method finds raw weights $w$ such that $S_\theta(w) + L_\theta(w) = W_{\mathrm{target}}$.
When these weights are then evaluated with the mismatch formula $\sign(w)(E_\theta(\abs{w}) + L_\theta(w))$, the behavior diverges by sign:
\begin{itemize}
    \item \textbf{Positive weights} ($w > 0$): Both formulas give the same result, since $S_\theta + L_\theta = \sign(w)(E_\theta + L_\theta)$ when $L_\theta > 0$.
    The initialization is correct.
    \item \textbf{Negative weights} ($w < 0$): The congruent formula gives $-E_\theta + L_\theta$ (symmetric-exponential opposes, linear reinforces), while the mismatch formula gives $-(E_\theta + L_\theta) = -(E_\theta - \abs{L_\theta})$ (both wrapped inside the negation, with the linear term's magnitude reducing the total).
    The effective weight magnitude is reduced by $2\abs{L_\theta}$ compared to the congruent target.
\end{itemize}

For our default parameters ($\beta = 7.5$), the linear pathway contributes roughly 10--15\% of the total effective weight magnitude for typical Xavier-scale weights. As beta increases, this further drops.
This means negative weights are under-initialized by approximately 20--25\% in magnitude, while positive weights are initialized correctly.
The result is an \textbf{asymmetric initialization} where positive weights start at full strength and negative weights start weaker.

We hypothesize the benefit comes primarily from symmetry breaking: the positive/negative asymmetry gives early gradients a preferred direction to act on, rather than requiring the network to break its own initial symmetry from a fully symmetric Xavier distribution.
This may be reinforced by our scheduled scale learning rates (\S\ref{sec:annealing}), which accelerate the learning rate of the scale parameters early on and may let the scales commit to a direction prior to the weights themselves beginning to converge, greater coupling their updates.

\paragraph{Xavier uniform over normal.}
We find that Xavier \emph{uniform} initialization consistently outperforms Xavier normal for the target weights before inversion.
We attribute this to the bounded support of the uniform distribution: extreme values from the tails of the normal distribution can produce very large raw parameters after inversion through the symmetric-exponential pathway, destabilizing early training.

\subsection{Structured Scale Parameters}
\label{sec:sparse_scale}

Each learnable scale ($e_w$, $l_w$, $m$, $s$, $n$) is parameterized with a chosen factorization pattern.
The optional residual $\Delta_\phi$ is an additive correction, and is usually parameterized as a LoRA-style update.
For a weight matrix $W \in \R^{d_{\mathrm{out}} \times d_{\mathrm{in}}}$, available patterns include:

\begin{itemize}
    \item \textbf{Global}: a single shared scalar.
    \item \textbf{Row / Column}: a vector along one axis ($d_{\mathrm{out}}$ or $d_{\mathrm{in}}$).
    \item \textbf{Row$\times$Col multiplicative}: two rank-1 vectors $r \in \R^{d_{\mathrm{out}}}$, $c \in \R^{d_{\mathrm{in}}}$ whose outer product $r \cdot c^\top$ gives the scale matrix. This is our default and recommended mode.
    \item \textbf{LoRA}: a low-rank additive perturbation $\text{base} + AB$ where $A \in \R^{d_{\mathrm{out}} \times r}$, $B \in \R^{r \times d_{\mathrm{in}}}$, initialized with $B = 0$~\citep{hu2021lora}.
\end{itemize}

Each scale has a \emph{learning rate multiplier} $k$ that scales its learning rate relative to the base rate, allowing different components of the reparameterization to learn at different speeds. One of the main reasons for this is that the scale parameters are initialized around 1.0 and have to travel a significant distance in weight space for a meaningful effect, as per the prior explanation of 0.5 as a halving operation and 2.0 as a doubling operation.

In our main runs, the output pathway scales $e_w$ and $l_w$ use \texttt{row\_col\_mult} as an efficient trade-off: it is more expressive than a global scalar while remaining much cheaper than full elementwise scale parameters.
The inner multiplier $m$ also uses \texttt{row\_col\_mult} in our experiments, though simplifying it to a row-only or column-only pattern is a natural future ablation.
For the offset $n$, we typically use a column-wise scale, since offset effects are more sensitive to input-axis structure.

\paragraph{Weight decay on scale parameters.}
We disable weight decay for scale parameters that primarily control the \emph{shape} of the reparameterization, such as the curvature multiplier $m$ and offset $n$.
However, we keep weight decay on the output pathway scales $e_w$ and $l_w$.
These scales directly weight the symmetric-exponential and linear contributions, so they can increase the overall effective weight magnitude even if the raw weight $w$ itself is decayed.
Applying weight decay to $e_w$ and $l_w$ prevents the model from growing the effective weight unboundedly by growing pathway scales.

\subsection{Exponential Pathway Learning Rate Annealing}
\label{sec:annealing}

The symmetric-exponential pathway has two controls that affect its strength: the outer scale $e_w$ and the inner multiplier $m$.
While both increase the exponential contribution, they serve different roles: $e_w$ can go negative (flipping the sign of the exponential part), while $m$ controls the curvature (how sharply the exponential grows with $\abs{w}$).

These controls are partially redundant, making joint optimization difficult.
We address this with a targeted learning rate annealing schedule: $e_w$ starts with a high $k$ value (fast learning) so it can quickly commit to a sign, then its learning rate decays over a cosine-in-log-space schedule, while $m$ starts with a low $k$ value and anneals upward to refine the curvature once $e_w$ has stabilized.

Specifically, for a parameter with initial multiplier $k_{\text{start}}$ and final multiplier $k_{\text{end}}$, the effective multiplier at step $t$ is:
\begin{equation}
k(t) = \exp\!\Bigl(\ln k_{\text{start}} + \frac{\ln(k_{\text{end}}/k_{\text{start}})}{2}\bigl(1 - \cos(\pi \cdot \min(t / T, 1))\bigr)\Bigr)
\end{equation}
where $T$ is the total annealing steps.
This interpolates between $k_{\text{start}}$ and $k_{\text{end}}$ in log space along a cosine curve, providing smooth transitions.
After step $T$, the multiplier remains at $k_{\text{end}}$.

\section{Experiments}
\label{sec:experiments}

We evaluate \methodshort on autoregressive language modeling. All experiments use the same codebase and training infrastructure.

\subsection{Experimental Setup}
\label{sec:setup}

\paragraph{Architecture.}
We use a modern transformer with GeGLU feedforward blocks~\citep{shazeer2020glu} (hidden dimension $\frac{8}{3}d$ rounded to the nearest multiple of 64), RoPE positional encodings~\citep{su2024roformer}, RMSNorm~\citep{zhang2019root}, and QK-normalization.
\methodshort is applied to all linear projections: Q, K, V, output projection, and both FFN layers.
Biases are enabled and also reparameterized.

\paragraph{Scale grid.}
We evaluate the full Cartesian product of widths \{1024, 2048, 3072\} and depths \{12, 24, 36\}, for nine total configurations.

\paragraph{Training.}
All models are trained on OpenWebText~\citep{gokaslan2019openwebtext} with sequence length 1024.
We use AdamW~\citep{loshchilov2017adamw} with $\beta_1 = 0.9$, $\beta_2 = 0.999$, weight decay 0.01, gradient clipping at 1.0, and linear learning rate warmup for 250 steps followed by linear decay.
Net batch size varies across runs; width-1024 configurations are trained for 250k steps, while width-2048 and width-3072 configurations are trained for 150k steps.
Models are compiled with \texttt{torch.compile} (reduce-overhead mode, full graph) and trained with bf16 mixed precision and gradient checkpointing. Compiling is important for minimizing the overhead of our method. Given that our method is composed entirely of elementwise operations, compiling offers a significant benefit in reducing the memory-bound parts of the computation.

\paragraph{\methodshort configuration.}
Unless otherwise noted, \methodshort experiments use:
the $\beta$ value selected by the initialization-scale heuristic described below,
mismatch forward mode with congruent initialization,
\texttt{row\_col\_mult} pattern for all scale parameters,
$k_{e_w} = 50$ (annealed to $k_{e_w,\text{end}} = 8.0$),
$k_m = 0.01$ (annealed to $k_{m,\text{end}} = 0.5$),
$k_{l_w} = 50$,(annealed to $k_{l_w,\text{end}} = 8.0$), 
optional LoRA-style residual $\Delta_\phi$ with $k_{\Delta} = 0.5$ when enabled,
weight decay enabled for $e_w$ and $l_w$ but disabled for the shape-controlling scale parameters,
Xavier uniform initialization,
fixed-$\beta$ curvature $\kappa=\beta m$ with $m$ initialized at one and the shared unnormalized normalizer $d=\beta$.

\subsection{Main Results}
\label{sec:main_results}

\begin{figure}[t]
\centering
\includegraphics[width=0.95\textwidth]{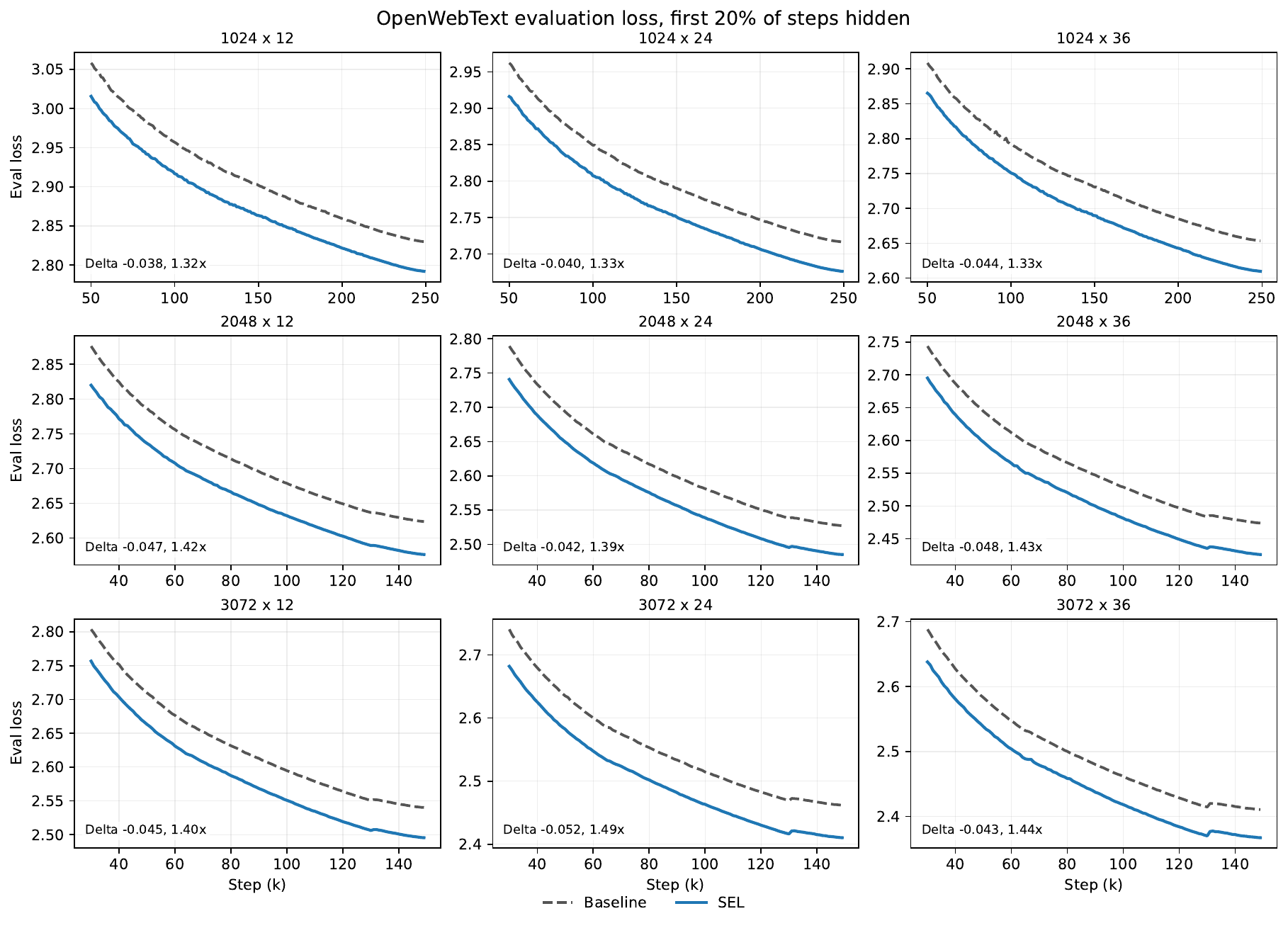}
\caption{\textbf{Evaluation loss during pretraining on OpenWebText} across all nine width $\times$ depth configurations. \emph{Top:} width 1024 (250k steps, $\beta = 7.5$). \emph{Middle:} width 2048 (150k steps). \emph{Bottom:} width 3072 (150k steps). $\beta$ is tuned per configuration at the wider scales. In each panel, the best \methodshort variant (solid) consistently reaches lower loss than the baseline (dashed). The annotation shows the final loss gap $\Delta$.}
\label{fig:main_results}
\end{figure}

\begin{table}[t]
\centering
\caption{\textbf{Main results on OpenWebText.} Best \methodshort variant vs.\ baseline at each scale. $\Delta$ Loss is \methodshort minus baseline (negative = better). Speedup is baseline steps divided by \methodshort steps to reach baseline's final loss e.g., 1.25$\times$ means \methodshort reaches the same loss in 25\% fewer steps. Width 1024 trains for 250k steps ($\beta = 7.5$); widths 2048 and 3072 train for 150k steps with per-depth $\beta$. $^\dagger$Best variant is mismatch \emph{without} offset (all others use mismatch + offset).}
\label{tab:main_results}
\begin{tabular}{lcccccc}
\toprule
Width & Depth & $\beta$ & Baseline & \methodshort & $\Delta$ & Speedup \\
\midrule
1024 & 12 & 7.5 & 2.830 & 2.792 & $-$0.038 & 1.32$\times$ \\
1024 & 24 & 7.5 & 2.717 & 2.676 & $-$0.040 & 1.33$\times$ \\
1024 & 36 & 7.5 & 2.654 & 2.610 & $-$0.044 & 1.33$\times$ \\
\midrule
2048 & 12 & 12.5 & 2.624 & 2.576 & $-$0.047 & 1.42$\times$ \\
2048 & 24 & 12 & 2.527 & 2.485 & $-$0.042 & 1.39$\times$ \\
2048 & 36$^\dagger$ & 15 & 2.474 & 2.426 & $-$0.048 & 1.43$\times$ \\
\midrule
3072 & 12$^\dagger$ & 17.5 & 2.540 & 2.495 & $-$0.045 & 1.40$\times$ \\
3072 & 24$^\dagger$ & 17.5 & 2.462 & 2.410 & $-$0.052 & 1.49$\times$ \\
3072 & 36$^\dagger$ & 20 & 2.411 & 2.367 & $-$0.043 & 1.44$\times$ \\
\bottomrule
\end{tabular}
\end{table}

Table~\ref{tab:variant_results} expands the main comparison to show the two \methodshort variants run at each scale: mismatch without the optional residual/offset, and mismatch with the optional LoRA-style residual plus offset.
At width 1024, the offset+LoRA variant is consistently strongest; at larger widths, the no-offset mismatch variant often matches or exceeds it.
We therefore report the better of the two in Table~\ref{tab:main_results}.

\begin{table}[t]
\centering
\caption{\textbf{Variant comparison on the main grid.} Final eval loss for baseline, mismatch, and mismatch with optional LoRA-style residual plus offset.}
\label{tab:variant_results}
\begin{tabular}{lcccc}
\toprule
Width & Depth & Baseline & Mismatch & Mismatch + offset/LoRA \\
\midrule
1024 & 12 & 2.830 & 2.799 & 2.792 \\
1024 & 24 & 2.717 & 2.687 & 2.676 \\
1024 & 36 & 2.654 & 2.637 & 2.610 \\
\midrule
2048 & 12 & 2.624 & 2.585 & 2.576 \\
2048 & 24 & 2.527 & 2.486 & 2.485 \\
2048 & 36 & 2.474 & 2.426 & 2.437 \\
\midrule
3072 & 12 & 2.540 & 2.495 & 2.512 \\
3072 & 24 & 2.462 & 2.410 & 2.416 \\
3072 & 36 & 2.411 & 2.367 & 2.372 \\
\bottomrule
\end{tabular}
\end{table}

Figure~\ref{fig:main_results} shows evaluation loss curves across all nine configurations spanning three widths and three depths.
\methodshort consistently outperforms the baseline throughout training at every scale (Table~\ref{tab:main_results}).
At width 1024 (250k steps), improvements grow with depth: $\Delta = -0.038$ at depth 12, $-0.040$ at depth 24, and $-0.044$ at depth 36.
At width 2048 (150k steps), the gains are larger: $\Delta = -0.047$, $-0.042$, and $-0.048$ for depths 12, 24, and 36.
At width 3072 (150k steps), gains remain strong: $\Delta = -0.045$, $-0.052$, and $-0.043$.
The largest single improvement is $\Delta = -0.052$ at 3072$\times$24.
Across all nine configurations, the baseline never reaches \methodshort's final loss within the same training budget.

\paragraph{Choosing $\beta$.}
We set $\beta$ using a simple initialization-scale heuristic rather than treating it as a broad sweep.
Because Xavier uniform initialization has bounded support that shrinks with matrix width, larger matrices require larger $\beta$ values for the symmetric-exponential curvature to become active near typical initial weight magnitudes.
In practice, we choose $\beta$ so that the exponential curve begins to noticeably depart from the linear regime around 1.5--2 initialization standard deviations, while remaining within the support of the Xavier uniform draw.
This gives $\beta = 7.5$ at width 1024, $\beta \in \{12.0, 12.5, 15.0\}$ at width 2048, and $\beta \in \{17.5, 20.0\}$ at width 3072.

\subsection{Ablation Studies}
\label{sec:ablations}

We explore initialization asymmetry, the exponential-only pathway, fixed and learned scale controls, structured scaling without curvature, optimizer-coordinate equivalence, the learning-rate schedule for scale parameters, and curvature parameterization.
The mechanism controls below form a separate, controlled 1024$\times$12 study: one seed (123), 100k optimizer steps, net batch 32, learning rate $3\times10^{-4}$, 2k warmup steps, $\beta=7.5$, and no offset or LoRA-style residual.
They use a common OpenWebText tokenization and evaluation stream, but are not presented as exact reproductions of the 250k-step width-1024 main-grid protocol.

\subsubsection{Initialization Asymmetry}

At width 1024, we run a comparison varying the combination rule (congruent vs mismatch) along with the optional offset/LoRA components (Table~\ref{tab:init_asymmetry_1024}).
The mismatch forward mode improves over congruent mode at every depth, and adding offset/LoRA further improves the mismatch runs at this scale.
The asymmetric sign-wrapping and initialization appears to be useful at all scales, though the main-grid results in Table~\ref{tab:variant_results} benefit from offset/LoRA changes at larger widths.

\begin{table}[t]
\centering
\caption{\textbf{Width-1024 initialization/combination comparison.} Final eval loss for the five runs available at width 1024. All \methodshort runs use congruent inversion; mismatch rows therefore include the initialization/forward asymmetry described in \S\ref{sec:mismatch}.}
\label{tab:init_asymmetry_1024}
\begin{tabular}{lccccc}
\toprule
Depth & Baseline & Congruent & Congruent + offset/LoRA & Mismatch & Mismatch + offset/LoRA \\
\midrule
12 & 2.830 & 2.819 & 2.832 & 2.799 & 2.792 \\
24 & 2.717 & 2.714 & 2.717 & 2.687 & 2.676 \\
36 & 2.654 & 2.645 & 2.649 & 2.637 & 2.610 \\
\bottomrule
\end{tabular}
\end{table}

The same pattern holds in the newly run 3072$\times$12 comparison at $\beta=20$ (Table~\ref{tab:init_asymmetry_3072}).
Mismatch improves over congruent mode both with and without the optional offset/LoRA branch, though in this wider setting the no-offset mismatch run is strongest.

\begin{table}[t]
\centering
\caption{\textbf{Width-3072 initialization/combination comparison.} Final eval loss at depth 12 with $\beta=20$.}
\label{tab:init_asymmetry_3072}
\begin{tabular}{lcccc}
\toprule
Width $\times$ depth & Congruent & Congruent + offset/LoRA & Mismatch & Mismatch + offset/LoRA \\
\midrule
3072$\times$12 & 2.770 & 2.801 & 2.741 & 2.762 \\
\bottomrule
\end{tabular}
\end{table}

\subsubsection{Exponential-Only Configuration}
\label{sec:exp_only}

A natural question is whether the linear pathway $L_\theta(w)$ is necessary for successful optimization, or whether the symmetric-exponential pathway alone is sufficient.
The exponential-only variant removes the linear pathway:
\begin{equation}
w_{\mathrm{eff}} = S_\theta(w) + \Delta_\phi
\end{equation}
omitting the $L_\theta(w)$ term from Eq.~\eqref{eq:full_transform}.

We train this configuration for the full 100k-step budget and vary whether the exponential output scale and curvature are learned (Table~\ref{tab:exp_only_controls}).

\begin{table}[t]
\centering
\caption{\textbf{Exponential-only ablation.} Final evaluation loss in the controlled 1024$\times$12 study. ``Fixed'' means that the corresponding auxiliary scale is exactly one.}
\label{tab:exp_only_controls}
\begin{tabular}{lccc}
\toprule
Variant & Exponential scale & Curvature & Final eval loss \\
\midrule
AdamW baseline & -- & -- & 2.9522 \\
Full SEL & learned & learned & \textbf{2.9295} \\
\midrule
Exp-only & fixed & fixed & 2.9471 \\
Exp-only & fixed & learned & 2.9456 \\
Exp-only & learned & fixed & 2.9433 \\
Exp-only & learned & learned & 2.9439 \\
\bottomrule
\end{tabular}
\end{table}

The exponential-only variants train normally and all finish modestly below the AdamW baseline.
Full SEL is better than the fully learned exponential-only arm at 199 of 200 evaluations after the first checkpoint, with a final gap of 0.0145 and roughly 4.5k fewer steps to reach eval loss 3.0.
This supports an advantage for the full configuration under this setup, but not a necessity claim: the exponential-only geometry is viable, and the comparison also changes initialization, combination mode, and the available scale variables.
Within the exponential-only $2\times2$ design, learning the exponential output scale gives a small final improvement, while the curvature effects are at milliloss scale and inconclusive from one seed.

\subsubsection{Frozen Geometry and Learned Scale Controls}
\label{sec:frozen_scale_controls}

To separate curved coordinates from auxiliary scale learning, we progressively freeze the curvature and pathway scales at one (Table~\ref{tab:frozen_scale_controls}).
Freezing a scale does not remove its pathway; only the exponential-only condition above removes $L_\theta$.

\begin{table}[t]
\centering
\caption{\textbf{Frozen and learned SEL controls.} All mismatch rows use congruent inversion. The final row uses congruent inversion and forward evaluation, providing matched effective Xavier initialization without learned scales.}
\label{tab:frozen_scale_controls}
\begin{tabular}{lccc}
\toprule
Variant & Pathway scales & Curvature & Final eval loss \\
\midrule
Full SEL & learned & learned & 2.9295 \\
SEL, fixed curvature & learned & fixed & \textbf{2.9177} \\
SEL, fixed pathway scales & fixed & learned & 2.9324 \\
SEL, all scales fixed (mismatch) & fixed & fixed & 2.9341 \\
SEL, all scales fixed (congruent) & fixed & fixed & 2.9445 \\
\bottomrule
\end{tabular}
\end{table}

All-fixed mismatch SEL remains 0.0181 below AdamW, and the congruent all-fixed control remains 0.0078 below AdamW.
Thus neither learned auxiliary degrees of freedom nor mismatched initialization alone explains the full effect.
The strongest configuration learns the linear and exponential pathway scales while holding curvature fixed, reaching 2.9177 and improving over all-fixed SEL by 0.0164.
By contrast, learning curvature with pathway scales fixed changes final loss by only 0.0018; adding the current learned-curvature schedule to already learned pathway scales is 0.0118 worse than fixing curvature.
This single-seed interaction motivates the curvature-optimization experiments in \S\ref{sec:curvature_param_ablation}; it does not establish that curvature adaptation is generally harmful.

\subsubsection{Structured-Linear Scale Controls}
\label{sec:structured_linear_controls}

To test whether learned multiplicative scales alone explain the SEL gains, we replace each ordinary projection by $W_{\mathrm{eff}}=A\odot W$ without exponential curvature and vary the structure of $A$.

\begin{table}[t]
\centering
\caption{\textbf{Ordinary structured-scale controls.} These models learn multiplicative scales around standard linear weights, without exponential curvature.}
\label{tab:structured_linear_controls}
\begin{tabular}{lc}
\toprule
Variant & Final eval loss \\
\midrule
AdamW baseline & 2.9522 \\
Linear scale, global & 2.9707 \\
Linear scale, row & 2.9495 \\
Linear scale, column & 2.9529 \\
Linear scale, row$\times$column & 2.9508 \\
\bottomrule
\end{tabular}
\end{table}

The ordinary row, column, and row$\times$column scale controls remain within 0.003 of AdamW, while the global scale is worse by 0.0185, so structured scaling alone does not reproduce the persistent SEL gains in this study.

\begin{figure*}[t]
\centering
\includegraphics[width=0.86\textwidth]{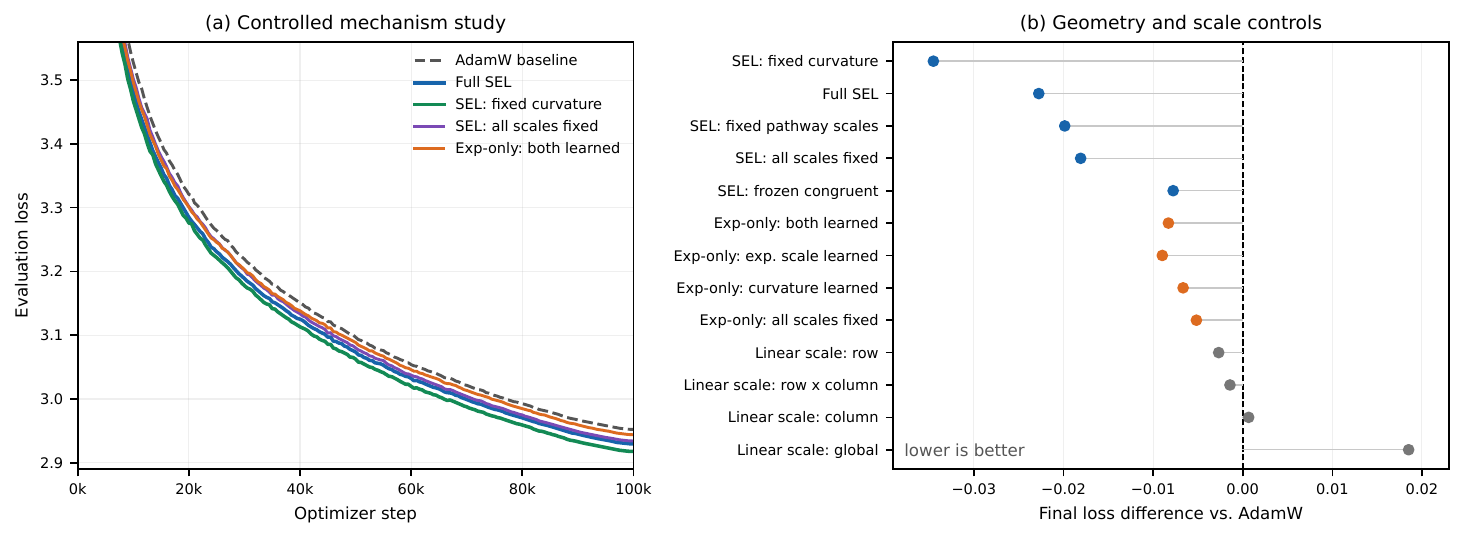}
\caption{\textbf{Controlled mechanism ablations at 1024$\times$12.}
\emph{Left:} evaluation trajectories for the AdamW baseline and representative SEL controls.
\emph{Right:} final loss relative to AdamW for frozen/learned SEL, exponential-only, and ordinary structured-scale controls; lower is better.
All curves use seed 123 and the shared 100k-step protocol.}
\label{fig:mechanism_ablations}
\end{figure*}

\subsubsection{Optimizer-Side Equivalence}
\label{sec:optimizer_equivalence}

A fixed invertible reparameterization can also be expressed as an optimizer-coordinate transformation.
We therefore compare frozen exponential-only SEL with AdamW against ordinary linear projections updated by \texttt{ExactExpOnlySELAdamW}.
The latter maps selected projection weights and biases into the same raw exponential coordinate for clipping, Adam moments, decoupled decay, and each update, while embeddings, norms, and the LM head retain ordinary AdamW updates.

\begin{table}[t]
\centering
\caption{\textbf{Optimizer-coordinate equivalence.} Final evaluation loss for the frozen parameterized transform and its optimizer-side implementation.}
\label{tab:optimizer_equivalence}
\begin{tabular}{lc}
\toprule
Variant & Final eval loss \\
\midrule
Frozen exp-only SEL + AdamW & 2.9471 \\
Linear + exact exp-only optimizer & 2.9461 \\
\bottomrule
\end{tabular}
\end{table}

The equivalence pair nearly overlaps across all 200 evaluations: mean absolute loss difference is $8.6\times10^{-4}$, the largest transient difference is $4.9\times10^{-3}$, and the final difference is $9.3\times10^{-4}$.
This is an implementation-level sanity check that frozen exponential-only SEL acts through optimizer coordinates rather than additional representational capacity.
It does not make adaptive learned geometry redundant: reproducing the jointly learned pathway and curvature scales optimizer-side would require carrying and optimizing equivalent additional state.

\begin{figure}[t]
\centering
\includegraphics[width=\textwidth]{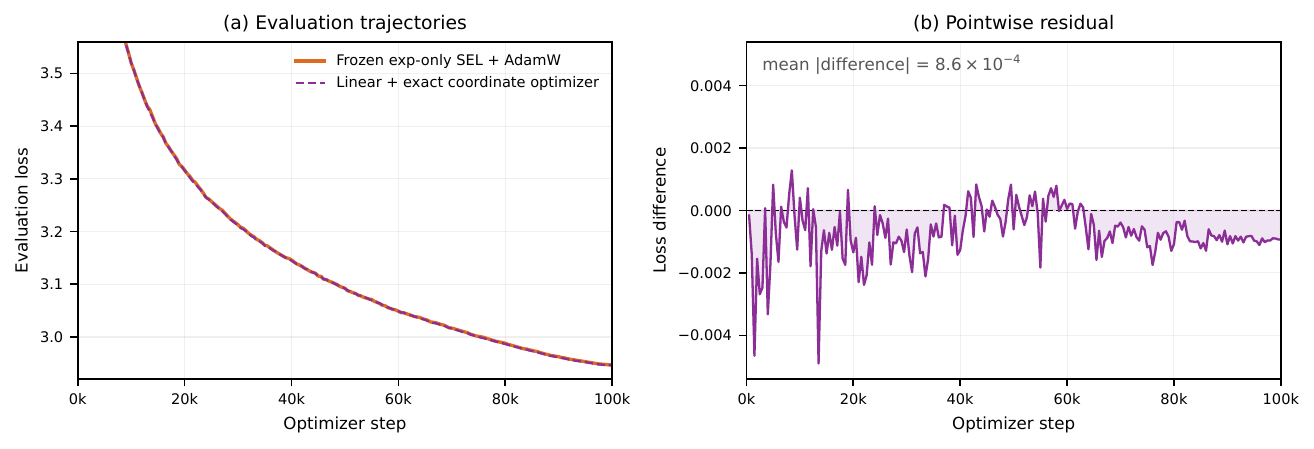}
\caption{\textbf{Frozen exponential-only SEL and its optimizer-coordinate implementation.}
\emph{Left:} evaluation-loss trajectories for the parameterized and optimizer-side implementations.
\emph{Right:} their pointwise loss difference, magnified by $10^3$.
The near-overlap holds across the complete 100k-step trajectory, rather than only at the final checkpoint.}
\label{fig:optimizer_equivalence}
\end{figure}

\subsubsection{LR Annealing}

The scale parameters may need to move quickly enough early in training to select a useful regime, then slow down once the transform has settled.
We therefore test whether the schedule in \S\ref{sec:annealing} matters by comparing the default schedule against fixed learning-rate multipliers.
The primary comparison is \textbf{default annealing} vs \textbf{fixed multipliers}.

\begin{figure}[t]
\centering
\begin{subfigure}{0.49\textwidth}
    \centering
    \includegraphics[width=\textwidth]{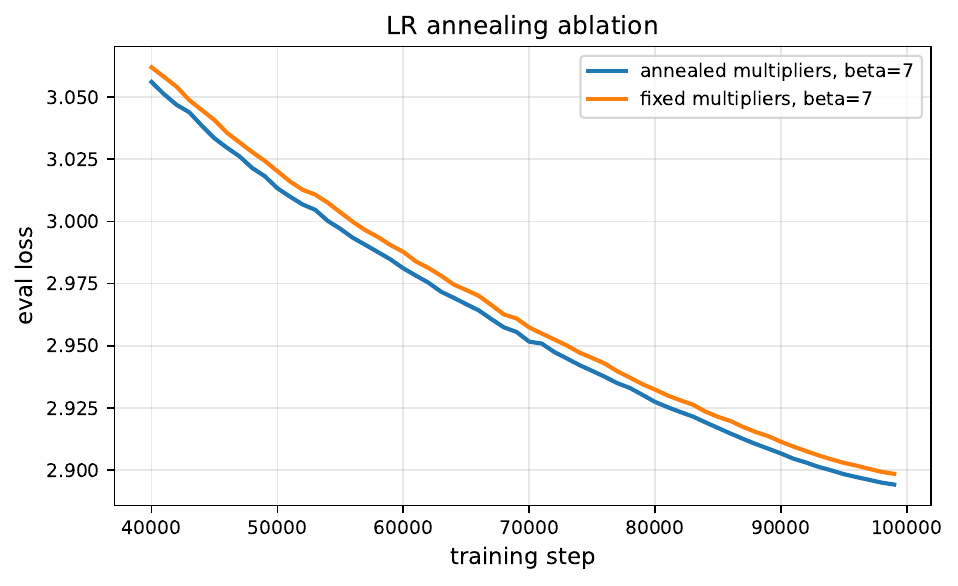}
\end{subfigure}
\begin{subfigure}{0.49\textwidth}
    \centering
    \includegraphics[width=\textwidth]{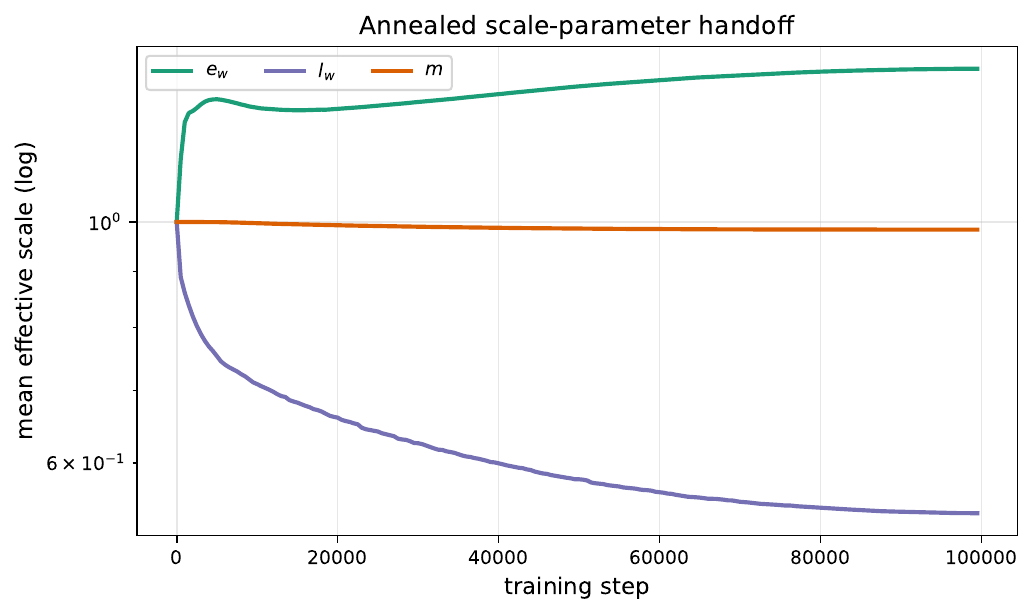}
\end{subfigure}
\caption{\textbf{LR annealing ablation and scale evolution.} \emph{Left:} eval loss for annealed vs.\ fixed scale multipliers, with the first 40\% of training omitted for readability. \emph{Right:} mean effective scales for $e_w$, $l_w$, and $m$ over training in the annealed run.}
\label{fig:lr_annealing_ablation}
\end{figure}

Table~\ref{tab:lr_annealing} and Figure~\ref{fig:lr_annealing_ablation} compare the default annealed schedule against fixed multipliers.
At the width-1024 curvature setting, the annealed run is slightly better at the end of training (2.894 vs.\ 2.899) with essentially identical runtime.
The scale trajectories show that most of the pathway rebalance occurs early: $e_w$ rises above its initialization, $l_w$ falls quickly, and $m$ remains close to one.

\begin{table}[t]
\centering
\caption{\textbf{LR annealing ablation at 1024$\times$12.} Both runs use mismatch mode, $\beta=7$, AdamW, and 100k training steps.}
\label{tab:lr_annealing}
\begin{tabular}{lc}
\toprule
Scale schedule & Final eval loss \\
\midrule
Annealed schedule & 2.894 \\
Fixed multipliers & 2.899 \\
\bottomrule
\end{tabular}
\end{table}

\subsubsection{Curvature Parameterization}
\label{sec:curvature_param_ablation}

Equations~\eqref{eq:curvature_normalizer}--\eqref{eq:exp_path} distinguish the exponential curvature coefficient $\kappa$ from the shared pathway normalizer $d$.
This raises two parameterization questions.
First, should the base curvature be treated as a fixed scalar, with $m$ initialized near one so that $\kappa=\beta m$, or should the learned inner multiplier absorb the base $\beta$ value directly so that $\kappa=\tilde{\beta}$?
The latter is more naturally a multiplicative quantity: moving from $\beta=7$ to $\beta=14$ and from $\beta=7$ to $\beta=3.5$ are symmetric changes in log space, but not in linear parameter space.
We therefore parameterize the absorbed-$\beta$ variant with a positive exponential scale, so additive raw updates map to multiplicative changes in the effective curvature of the exponential pathway.

Second, should changes in $\kappa$ also change the shared pathway normalizer?
The unnormalized variants use $d=\beta$, whereas \texttt{normalize\_by\_inner\_mult} uses $d=\kappa$.
The latter exactly preserves the exponential path's local slope, $e_w\exp(-n)$, as $\kappa$ changes.
It also reduces the direct linear contribution to $l_w/\kappa$, so at the default $e_w=l_w=1$, $n=0$, and practical values $\kappa\geq7.5$, the total near-zero gain remains within $13.3\%$ of the exponential path's unit gain while the curvature rate changes.
We evaluate the resulting $2\times2$ design in a 100k-step ablation (Table~\ref{tab:curvature_param_ablation}).

\begin{table}[t]
\centering
\caption{\textbf{Curvature parameterization ablation.} Actual 1024$\times$12, 100k-step comparison of fixed $\kappa=\beta m$ versus directly learned $\kappa=\tilde{\beta}$, and of the unnormalized choice $d=\beta$ versus effective-curvature normalization $d=\kappa$. Lower eval loss is better.}
\label{tab:curvature_param_ablation}
\begin{tabular}{lc}
\toprule
Variant & Final eval loss \\
\midrule
Fixed $\beta$, unnormalized & 2.899 \\
Learned/log-space $\tilde{\beta}$, unnormalized & 2.893 \\
Fixed $\beta$, normalized & 2.895 \\
Learned/log-space $\tilde{\beta}$, normalized & \textbf{2.888} \\
\bottomrule
\end{tabular}
\end{table}

The combined variant performs best in this sweep, suggesting that log space curvature updates and effective curvature normalization are complementary rather than redundant. For future work, it may be worth repeating the main experiments with the combined variant.

\subsection{Analysis}
\label{sec:analysis}

\subsubsection{Weight Distribution Under the Transform}

\begin{figure}[t]
\centering
\includegraphics[width=0.9\textwidth]{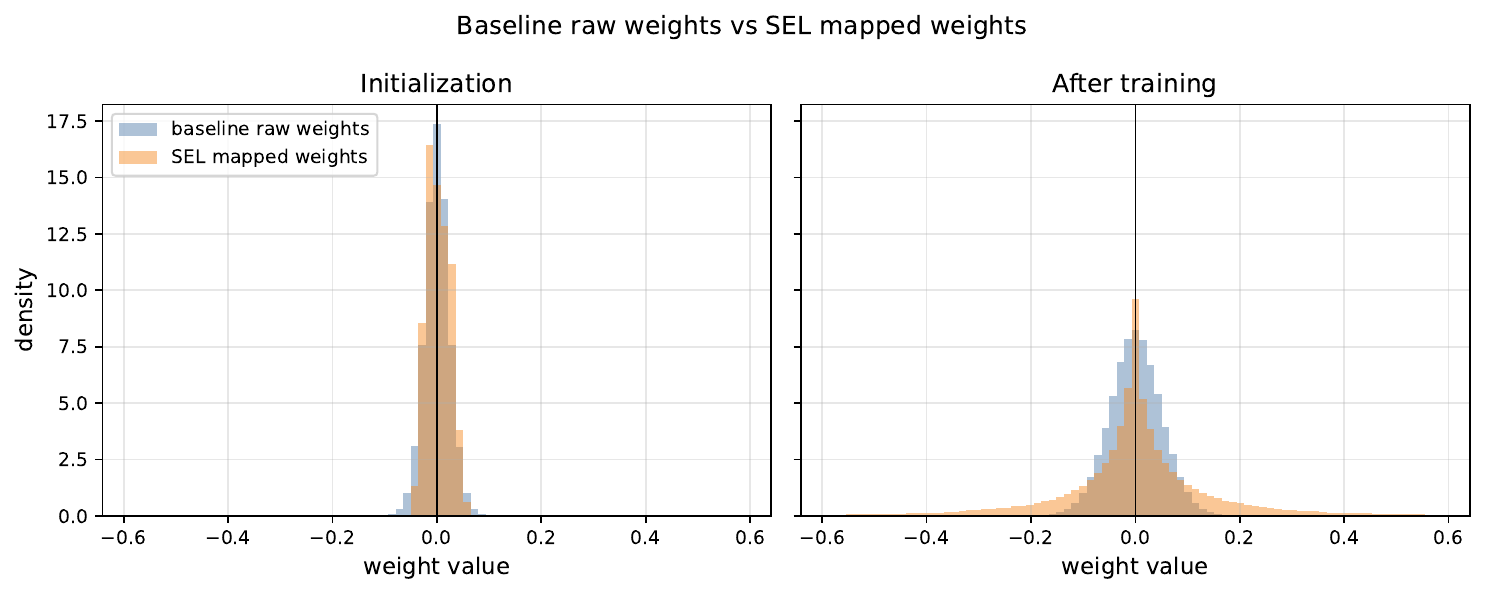}
\caption{\textbf{Baseline vs.\ \methodshort weight distributions.} Distribution of weights in a regular baseline parameterization training run (blue) vs. and SEL training run at dim=1024 and depth=12. Left shows the weight distribution at initialization and right shows the weight distribution at the end of 100k training steps.}
\label{fig:raw_effective_hist}
\end{figure}

Figure~\ref{fig:raw_effective_hist} compares standard baseline weights against mapped \methodshort weights at initialization and after training for matched 1024$\times$12 runs.
At initialization, the baseline raw weights and \methodshort effective weights are close by construction.
After training, the \methodshort transform produces a visibly different, more heavy-tailed effective-weight distribution than the baseline parameterization.

\subsubsection{Scale Parameter Evolution}

The scale trajectories in Figure~\ref{fig:lr_annealing_ablation} show the change in $e_w$, $l_w$, and $m$ throughout training.
In the annealed run, averaged across tracked effective-scale tensors, $e_w$ rises from 1.0 to about 1.38, while $l_w$ falls from 1.0 to about 0.54 by the final evaluation.
The inner multiplier $m$ changes much less, ending near 0.98.
Most of the visible pathway rebalance occurs early in training, especially the initial rise in $e_w$ and drop in $l_w$.

\subsubsection{Transform Shape Visualization}

\begin{figure}[t]
\centering
\includegraphics[width=0.95\textwidth]{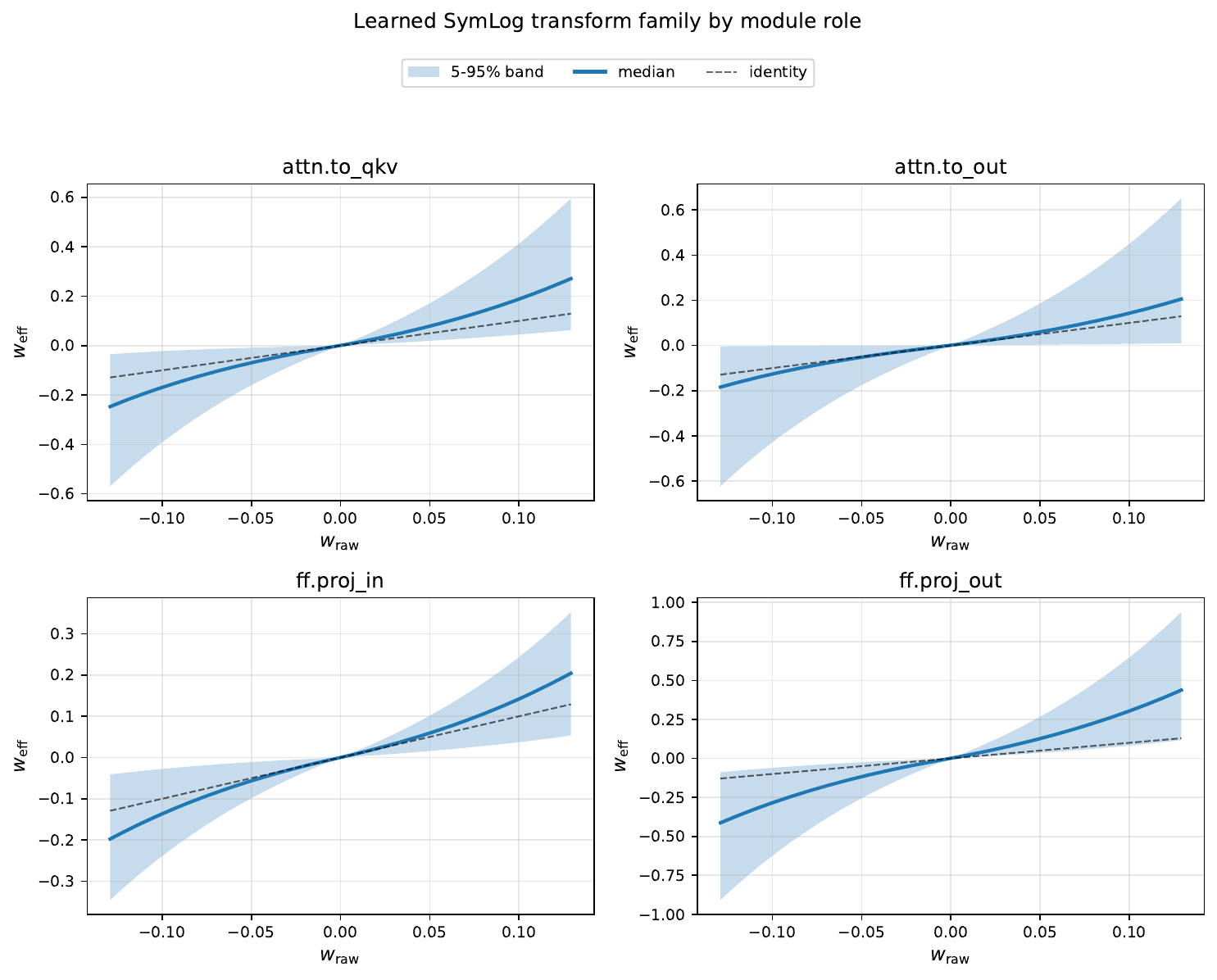}
\caption{\textbf{Learned transform shapes by module role.} For sampled weight positions in the 1024$\times$12 annealed run with $\beta=7.5$, we evaluate $w_{\mathrm{eff}}$ as a function of $w_{\mathrm{raw}}$ using the learned per-position scales. Shaded bands show the 5--95\% range across sampled positions; dashed lines show the identity. The learned transforms preserve small-weight control near zero while amplifying larger raw magnitudes.}
\label{fig:transform_family}
\end{figure}

Figure~\ref{fig:transform_family} plots the learned transform by module role for the $\beta=7.5$ analysis run. We observe in many cases the median example has a tilt, induced by modification of scale parameters, in slope at origin, pointing slightly upwards.
The visualization supports our hypothesis that the symmetric-exponential shape is utilized, or even made more extreme, evidenced by the downweighting of the linear path while upweighting of the exponential path.

\begin{figure}[t]
\centering
\includegraphics[width=0.95\textwidth]{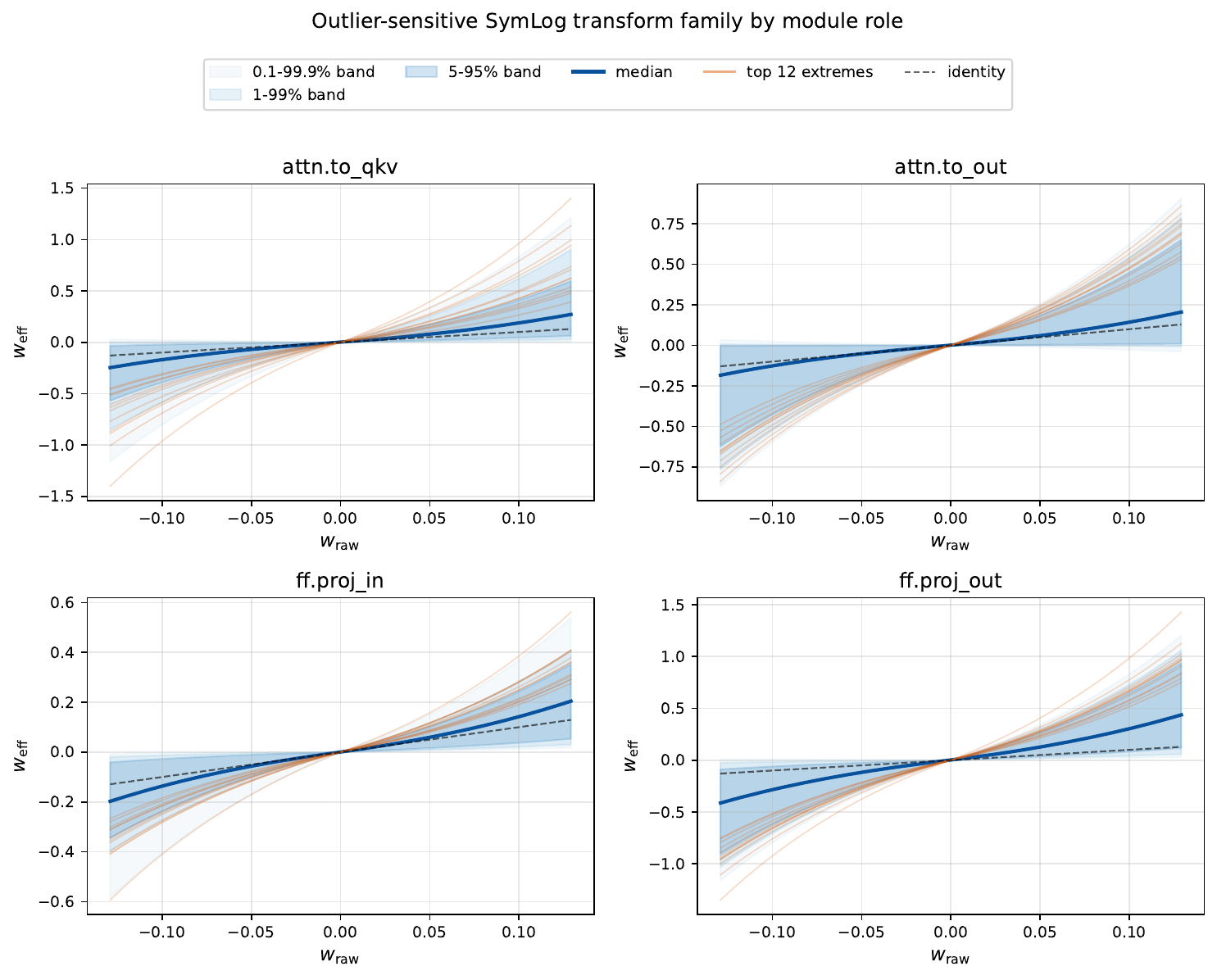}
\caption{\textbf{Outlier-sensitive transform shapes by module role.} The same learned transforms as Figure~\ref{fig:transform_family}, but with wider 1--99\% and 0.1--99.9\% bands plus individual curves from the largest sampled raw/effective weight positions. Rare positions show much stronger tail amplification than the median, matching the heavy-tail motivation for the parameterization.}
\label{fig:transform_outlier_bands}
\end{figure}

\section{Discussion}

\paragraph{Why symmetric-exponential + linear?}
The symmetric-exponential pathway provides the magnitude-dependent sensitivity and multiplicative updates: small weights remain in a near-linear regime, while larger raw weights enter a curved regime where linear raw-parameter update steps can produce a larger effective-weight change.
The linear pathway gives the model a direct additive route from raw to effective weights, and the learned pathway scales allow training to choose how much to rely on each branch.
This is loosely analogous to a residual path in the limited sense that gradients need not pass only through the curved branch, though both pathways are functions of the same underlying weight.
The exponential-only ablation in \S\ref{sec:exp_only} shows that the curved branch alone trains successfully and modestly improves over AdamW; the full configuration performs better under the controlled setup, but the linear pathway is not necessary for learning.

Viewed through the lens of the optimizer-initialization mismatch described in Section~\ref{sec:motivation}, \methodshort can be understood as making output space parameter updates proportional to weight magnitude.
Under standard parameterization with Adam, a weight at magnitude $0.01$ and one at magnitude $1.0$ receive updates of similar absolute size, but the small weight is perturbed by $100\times$ more in relative terms.
\methodshort's symmetric-exponential pathway counteracts this by increasing $\partial w_{\mathrm{eff}} / \partial w_{\mathrm{raw}}$ with $\abs{w}$: a comparable raw-space step corresponds to a larger movement in effective weight space once the parameter has entered the curved regime.
Thus the parameterization offers the relative scaling that a standard parameterization with Adam would not have, without requiring per-parameter learning rate tuning.
The optimizer-equivalence control in \S\ref{sec:optimizer_equivalence} makes this interpretation concrete for frozen exponential-only SEL: explicitly applying AdamW in the corresponding raw coordinate closely reproduces the parameterized trajectory.
Learned SEL scales extend this fixed coordinate change into an adaptive, jointly optimized geometry.
$\mu$P~\citep{yang2022tensor} addresses a related but separate axis of the same mismatch: it rescales per-tensor learning rates from static, width-dependent init statistics/typical dimension magnitudes, whereas \methodshort adapts effective step size per-parameter on the fly as weights grow. The two are complementary and freely composable.

\paragraph{Mismatch Form.}
A surprising finding in our ablations is that mismatched initialization outperforms the congruent versions.
The asymmetry introduced by the mismatch (weaker negative weights) may provide a useful inductive bias or early symmetry-breaking effect.

\paragraph{Structured scales as lightweight adaptation.}
The structured scale parameters add a small number of auxiliary parameters (two vectors of size $d_{\mathrm{out}}$ and $d_{\mathrm{in}}$ per scale, per layer) that modulate the reparameterization.
This is significantly cheaper than full element-wise scales and provides enough expressivity to adapt the transform per-row and per-column.
The \texttt{row\_col\_mult} pattern is balanced between expressivity and cost for the output pathway scales. Each output neuron and each input feature has its own multiplicative adjustment to the exponential/linear balance.
We also experimented with additive row+column composition, which is appealing for quantities that behave like log-space shifts, such as the inner multiplier or offset.
However, additive row+column scales introduce a gauge symmetry: adding a constant to the row vector and subtracting it from the column vector leaves the combined scale unchanged.
In practice this underperformed the multiplicative row$\times$column form, so we use \texttt{row\_col\_mult} by default.
The controlled freeze study refines this motivation: pathway-scale learning provides the clearest additional gain, while fixed curved geometry already improves over AdamW and the present learned-curvature schedule does not improve on fixed curvature.

\paragraph{Future work: pretrained conversion and finetuning.}
Because \methodshort changes only the parameterization of weights, conversion is possible in both directions.
After training, effective weights can be materialized and deployed as ordinary linear layers.
Conversely, a standard pretrained weight matrix can be inverted into the \methodshort raw parameter space by solving $f(W_{\mathrm{raw}}) = W_{\mathrm{pre}}$, yielding an equivalent model at the moment of conversion.
This makes continued pretraining and finetuning in \methodshort space a natural direction for future work.

\paragraph{Training and inference cost.}
A key practical advantage of \methodshort over architectural modifications (e.g., mixture-of-experts, nonlinear branches) is that it changes the parameterization without changing the deployed architecture.
During training, the transform consists of element-wise operations on weight tensors, so compiled implementations using \texttt{torch.compile}, Triton kernels, and fused AdamW keep the overhead modest.
Table~\ref{tab:profiling} shows preliminary H200 profiling for representative small and large configurations.
Forward/backward overhead is small, while the fused AdamW step shows the most visible increase.
Combining per-step overhead with the step reductions in Table~\ref{tab:main_results}, \methodshort still yields estimated wall-clock speedups to matched validation loss.

\begin{table}[t]
\centering
\caption{\textbf{Preliminary H200 profiling and estimated wall-clock speedup.} Timings are milliseconds per training step component, reported as baseline/\methodshort. Step overhead is computed from the summed forward, backward, and optimizer times. Estimated wall-clock speedup divides the step reduction from Table~\ref{tab:main_results} by the measured per-step overhead.}
\label{tab:profiling}
\begin{tabular}{lcccccc}
\toprule
Config & Fwd & Bwd & Opt & Step overhead & Step speedup & Wall-clock speedup \\
\midrule
1024$\times$12 & 67/68 & 214/209 & 6/15 & +1.7\% & 1.32$\times$ & 1.29$\times$ \\
3072$\times$36 & 265/271 & 895/940 & 46/62 & +5.5\% & 1.44$\times$ & 1.37$\times$ \\
\bottomrule
\end{tabular}
\end{table}

These measurements indicate that the transform's training cost is small relative to the reduction in steps.
After training, one computes $W_{\mathrm{eff}} = f(W_{\mathrm{raw}})$ once and converts to standard linear layers.
All auxiliary parameters (structured scales, the curvature $\beta$, etc.)\ are folded into the effective weights and discarded, giving \textbf{zero additional inference cost}.

\paragraph{Alternative formulations: sinh.}
The hyperbolic sine function $\sinh(x) = (e^x - e^{-x})/2$ shares many properties with symexp: it is an odd function that behaves approximately linearly for small $\abs{x}$ (since $\sinh(x) \approx x$ for $\abs{x} \ll 1$) and grows exponentially for large $\abs{x}$.
A key distinction is that sinh is infinitely differentiable, whereas symexp is continuously differentiable but not twice differentiable at the origin.
A sinh-based reparameterization $w_{\mathrm{eff}} = \sinh(\beta w) / \beta$ would provide similar magnitude-dependent curvature with smooth derivatives of all orders.
We leave systematic comparison of sinh-based transforms to future work, noting that preliminary experiments suggest comparable performance with potentially improved optimization smoothness.

\paragraph{Interaction with weight decay.}
Weight decay is itself magnitude-aware: applying $w \leftarrow w - \lambda w = w(1 - \lambda)$ shrinks parameters in proportion to their current value, so larger parameters receive larger absolute shrinkage.
\methodshort introduces a different kind of magnitude dependence through the raw-to-effective map: as $\abs{w_{\mathrm{raw}}}$ grows, the local sensitivity $\partial w_{\mathrm{eff}} / \partial w_{\mathrm{raw}}$ increases, so comparable raw-space steps can produce larger movements in effective weight space.
Thus both mechanisms act in a magnitude-aware manner.

\paragraph{Weight decay on auxiliary scales.}
Decaying the raw weights alone does not constrain every route by which effective weights can grow: the output pathway scales $e_w$ and $l_w$ directly multiply the symmetric-exponential and linear contributions. Thus, they can increase effective magnitudes even when $w_{\mathrm{raw}}$ is decayed.
We therefore apply weight decay to $e_w$ and $l_w$, while disabling it for geometry-shaping parameters such as the inner multiplier $m$ and offset $n$.

\paragraph{Outliers at scale and their functional importance.}
Recent work has shown that larger language models develop increasingly pronounced weight and activation outliers. These outliers are not demonstrated to be critical to model performance and degrade performance if removed.
\citet{dettmers2022llm} documented that outlier features emerge at scale. Models under $\sim$6B parameters have few outliers, but larger models develop specific dimensions with extreme values that dominate attention.
\citet{sun2024massive} found ``massive activations'' in LLMs that can be 100,000$\times$ larger than typical values; these activations remain constant regardless of input and function as indispensable bias terms.
Most strikingly, \citet{yu2024superweight} showed that pruning a \emph{single} ``super weight'' can destroy an LLM's ability to generate coherent text, increasing perplexity by three orders of magnitude.
These super weights, typically found in \texttt{mlp.down\_proj} of early layers, create persistent super activations that suppress stopword likelihoods in final logits.

This body of work suggests that during training, some parameters must grow to extreme values and that these are the parameters most disadvantaged by linear additive updates.
\methodshort's exponential amplification directly addresses this: as weights grow, their effective gradients are amplified, accelerating progress toward large-magnitude targets.
At scale, where initialization values shrink further (due to larger matrix dimensions) while the need for extreme outliers persists or intensifies, larger $\beta$ values may be needed to provide greater curvature and accelerated update speed to reach final values.

\paragraph{Multiplicative computation, multiplicative parameterization.}
A useful intuition for \methodshort comes from considering how parameters affect computation.
LayerNorm scale parameters provide a clean example: initialized at $\gamma = 1.0$, moving to $\gamma = 0.5$ halves the output scale while moving to $\gamma = 2.0$ doubles it.
These are symmetric operations in terms of their \emph{computational effect} ($\div 2$ vs $\times 2$), yet asymmetric in optimization distance ($0.5$ vs $1.0$).
Under Adam's near-constant update magnitudes, ``doubling'' takes twice as many steps as ``halving.''
The natural parameterization for such multiplicative quantities is logarithmic: with $\gamma_{\mathrm{eff}} = \exp(\gamma_{\mathrm{raw}})$, both operations require $\abs{\Delta \gamma_{\mathrm{raw}}} = \log 2$.

We suggest this intuition extends, at least partially, to matrix weights.
In $y = Wx$, each weight $w_{ij}$ multiplicatively scales $x_j$'s contribution to $y_i$.
A weight growing from $0.01 \to 0.1$ and one growing from $0.1 \to 1.0$ both represent $10\times$ multiplicative changes, but the latter requires covering $10\times$ more optimization distance.
Not all aspects of the function implemented by the composed neural network may act in a relative manner: biases are additive, activation functions have thresholds at absolute values.
However, \methodshort attempts to respect both regimes.
The offset $n$ provides explicit control over this transition.
This ``soft log-space'' parameterization may explain why the symmetric-exponential pathway helps weights reach large-magnitude targets that the network's computation requires.

\section{Limitations}

\begin{itemize}
    \item \textbf{Implementation dependence}: The training transform is element-wise and is applied on every forward pass, so its overhead depends on how well these operations are fused by \texttt{torch.compile}/Triton or similar compilers. The additional scale parameters also increase the number of optimizer tensors; foreach or fused optimizer implementations reduce the overhead compared with Python loops over many small tensors.
    \item \textbf{Hyperparameter sensitivity}: The method introduces several hyperparameters ($\beta$, $k$ values for each scale, annealing schedule). We use \texttt{row\_col\_mult} as a robust structured-scale default and choose $\beta$ with the width-dependent initialization heuristic above, but further work is needed to understand how these choices should scale beyond our tested model sizes.
    \item \textbf{Scale}: Our largest configuration (3072$\times$36) is relatively small by modern standards. Validation at larger scales would strengthen the results.
    \item \textbf{Task diversity}: We evaluate only autoregressive language modeling. The method may behave differently on other tasks (classification, generation with different loss functions, multimodal models).
    \item \textbf{Theoretical understanding}: We provide intuitive explanations for why mismatched initialization helps but lack formal analysis. A rigorous understanding of the optimization landscape under this reparameterization remains open.
    \item \textbf{Mechanism-study replication}: The exponential-only, frozen-scale, structured-linear, and optimizer-equivalence controls use one seed in a separate 100k-step 1024$\times$12 protocol. Persistent differences motivate the mechanistic interpretation, but milliloss-scale rankings and curvature-scale interactions require replication.
\end{itemize}

\section{Conclusion}

We introduced \method (\methodshort), a nonlinear weight reparameterization that decomposes transformer linear layer weights into symmetric-exponential and linear components.
One useful insight is that a deliberate asymmetry between initialization and forward-pass combination rules can improve optimization in small-scale ablations, possibly by acting as early symmetry breaking.
Learnable scale parameters allow modifying the the weight overall transform, and a targeted annealing schedule manages the transition between partially redundant exponential controls.
\methodshort has modest training overhead in compiled implementations because its transform is element-wise, and adds no inference cost because it requires only a one-time weight materialization after training.

Signed exponential-only SEL trains normally, so the linear pathway is not necessary for learning under this setup.
More broadly, fixed curved geometry improves over AdamW, learned pathway scales can provide a further gain, and an exact optimizer-coordinate implementation reproduces frozen exponential-only SEL.
Together with the main grid, these results support curved weight-space optimization as the central effect while identifying adaptive scale learning as a useful but separable extension.

\bibliography{references}

\appendix

\section{Implementation Details}
\label{app:implementation}

\paragraph{\method Transform (Full).}
The complete forward pass for a weight matrix $W \in \R^{d_{\mathrm{out}} \times d_{\mathrm{in}}}$ is:
\begin{enumerate}
    \item Compute effective curvature: $\kappa = \beta \cdot m()$ in the fixed-$\beta$ form, or $\kappa=\tilde{\beta}()$ in the absorbed-$\beta$ form.
    \item Select the shared normalizer: $d=\beta$ for the unnormalized form or $d=\kappa$ for effective-curvature normalization.
    \item Compute exponential part: $E = \frac{e_w()}{d}\bigl(\exp(\kappa\abs{w} - n()) - \exp(-n())\bigr)$, where $n()$ is the offset and $e_w()$ is the relative exponential scale.
    \item Compute linear part: $L = \frac{l_w()}{d}w$, where $l_w()$ is the relative linear scale.
    \item Combine (mismatch mode): $w_{\mathrm{eff}} = \sign(w)(E + L) + \Delta_\phi$.
\end{enumerate}

\paragraph{Newton's Method Initialization.}
Given target weights $W_{\mathrm{target}}$ from Xavier uniform initialization, we solve for raw weights via $n_{\mathrm{iter}} = 10$ iterations of the congruent formula (shown for the default $n=0$):
\begin{align}
    f(x) &= \sign(x) \cdot \frac{e_{w,0}}{d_0}\bigl(\exp(\kappa_0\abs{x}) - 1\bigr) + \frac{l_{w,0}}{d_0}x \\
    f'(x) &= e_{w,0}\frac{\kappa_0}{d_0} \cdot \exp(\kappa_0\abs{x}) + \frac{l_{w,0}}{d_0} \\
    x_{k+1} &= x_k - \frac{f(x_k) - W_{\mathrm{target}}}{f'(x_k)}
\end{align}
Here $e_{w,0}$ and $l_{w,0}$ are the initial relative pathway scales, both set to one. In the canonical base form, we have $\kappa_0=d_0=\beta$.
The iteration converges rapidly (typically within 5 or 6 iterations) since $f$ is monotonically increasing and continuously differentiable, with a discontinuous second derivative at $x=0$.

\paragraph{Structured Scale Modes.}
The \texttt{row\_col\_mult} mode stores two vectors $r \in \R^{d_{\mathrm{out}}}$ and $c \in \R^{d_{\mathrm{in}}}$, initialized so that $r_i \cdot c_j = v$ for all $i, j$ where $v$ is the target value.
The effective scale is $r \cdot c^\top$, broadcast to match the weight matrix shape.
For multiplicative composition, each component is initialized to $\sqrt{v}$ (or $v^{1/2}$ from the inverse transform if a nonlinear transform is applied).

\section{Alternative Formulations}
\label{app:formulations}

During development, we explored several alternative formulations before arriving at the version presented in the main paper.
Here we document these variants for completeness.

\subsection{Combination Modes}

The main paper focuses on the \texttt{congruent} and \texttt{mismatch} modes for combining the symmetric-exponential and linear pathways.
Two additional modes were explored:

\paragraph{Oppose mode.}
In this variant, the linear pathway \emph{subtracts} from the exponential:
\begin{equation}
    w_{\mathrm{eff}} = S_\theta(w) - L_\theta(w)
\end{equation}
where $S_\theta(w)$ is the symmetric-exponential pathway and $L_\theta(w) = w \cdot l_w$ is the linear pathway.
This creates an asymmetry where the symmetric-exponential pathway must ``overcome'' the linear opposition.
We observed reduced performance in this configuration, possibly because the opposing forces create optimization challenges.

\paragraph{Exp-only mode.}
This variant removes the linear pathway entirely: $w_{\mathrm{eff}} = S_\theta(w)$.
Without the linear pathway, the transform behaves purely exponentially for all weight magnitudes.
In the controlled study, this variant trained normally and modestly improved over AdamW, although the full configuration reached a lower loss.

\subsection{Damped Exponential Formulation}

An alternative formulation introduces explicit damping to prevent exponential blowup for large weight magnitudes:
\begin{equation}
    w_{\mathrm{eff}} = \sign(w) \cdot \frac{\exp(\beta\abs{w} - n) - \exp(-n)}{\beta \cdot \exp(\abs{w})} + w \cdot \left(1 - \exp(-n \cdot \exp(\abs{w}))\right)
\end{equation}

This formulation has two key modifications:

\paragraph{Damping denominator.}
The symmetric-exponential pathway is divided by $\beta \cdot \exp(\abs{w})$ rather than just $\beta$.
For large $\abs{w}$, this grows as $\exp(\abs{w})$, which partially cancels the $\exp(\beta\abs{w})$ in the numerator.
The net effect is that the symmetric-exponential pathway grows as $\exp((\beta - 1)\abs{w})$ for large weights, taming the otherwise unbounded exponential growth when $\beta > 1$.

\paragraph{Input-dependent linear coefficient.}
Instead of a constant linear pathway $w / \beta$, the coefficient $(1 - \exp(-n \cdot \exp(\abs{w})))$ depends on the weight magnitude.
For small $\abs{w}$, this coefficient is small (suppressing the linear term), while for large $\abs{w}$ it rapidly saturates to 1.
This gives the linear pathway a naturally high contribution for weights that have already grown large, providing stability in the large-weight regime.

The damped formulation is more conservative than the version in the main paper: it provides magnitude-dependent curvature while explicitly preventing runaway growth.
However, in practice we found that the simpler undamped formulation (with appropriate $\beta$ tuning) achieved comparable or better results with fewer hyperparameters to tune.

\section{Supplementary Figures}
\label{app:figures}

\subsection{Transform Visualizations}

\begin{figure}[h]
\centering
\includegraphics[width=0.7\textwidth]{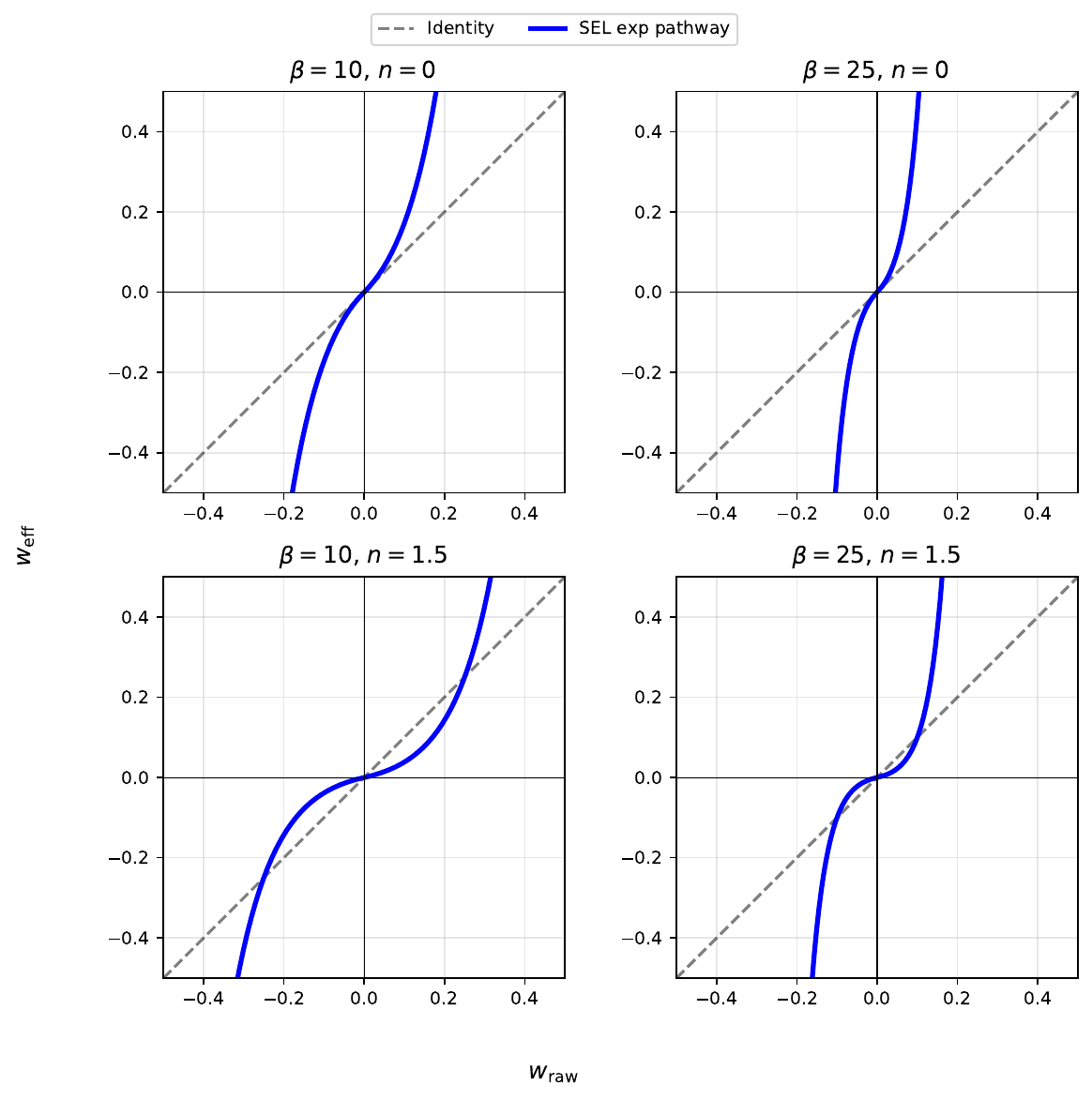}
\caption{\textbf{Combined effect of $\beta$ and offset $n$.} A 2$\times$2 grid showing how both parameters jointly affect the transform shape. Higher $\beta$ increases curvature; higher $n$ suppresses the exponential near zero.}
\label{fig:beta_offset_grid}
\end{figure}

\begin{figure}[h]
\centering
\includegraphics[width=0.85\textwidth]{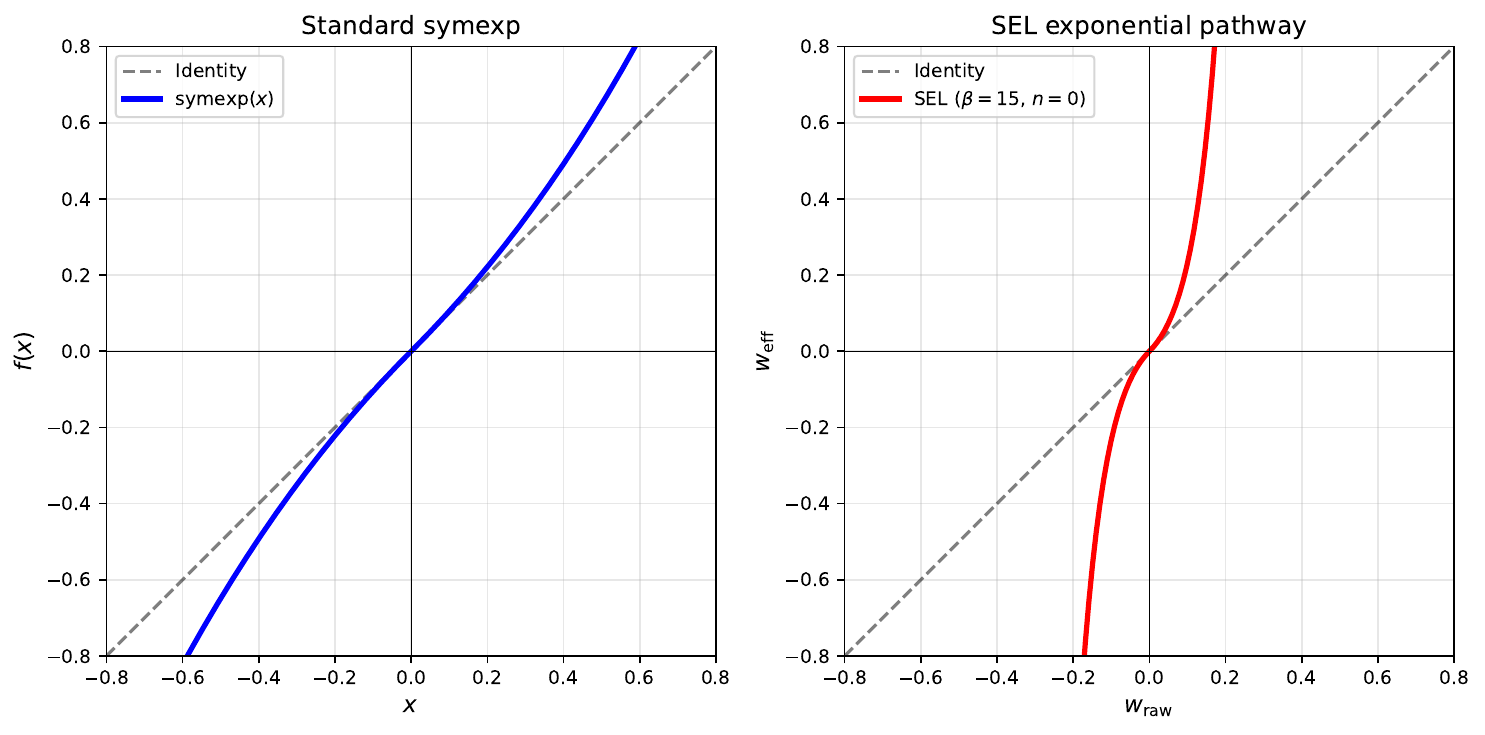}
\caption{\textbf{Standard symexp vs.\ SEL symmetric-exponential pathway.} \emph{Left:} The standard symexp function has fixed curvature. \emph{Right:} SEL with $\beta=15$ provides controllable curvature while maintaining similar qualitative behavior.}
\label{fig:symexp_vs_sel}
\end{figure}

\subsection{Mismatched Initialization Analysis}

\begin{figure}[h]
\centering
\includegraphics[width=0.95\textwidth]{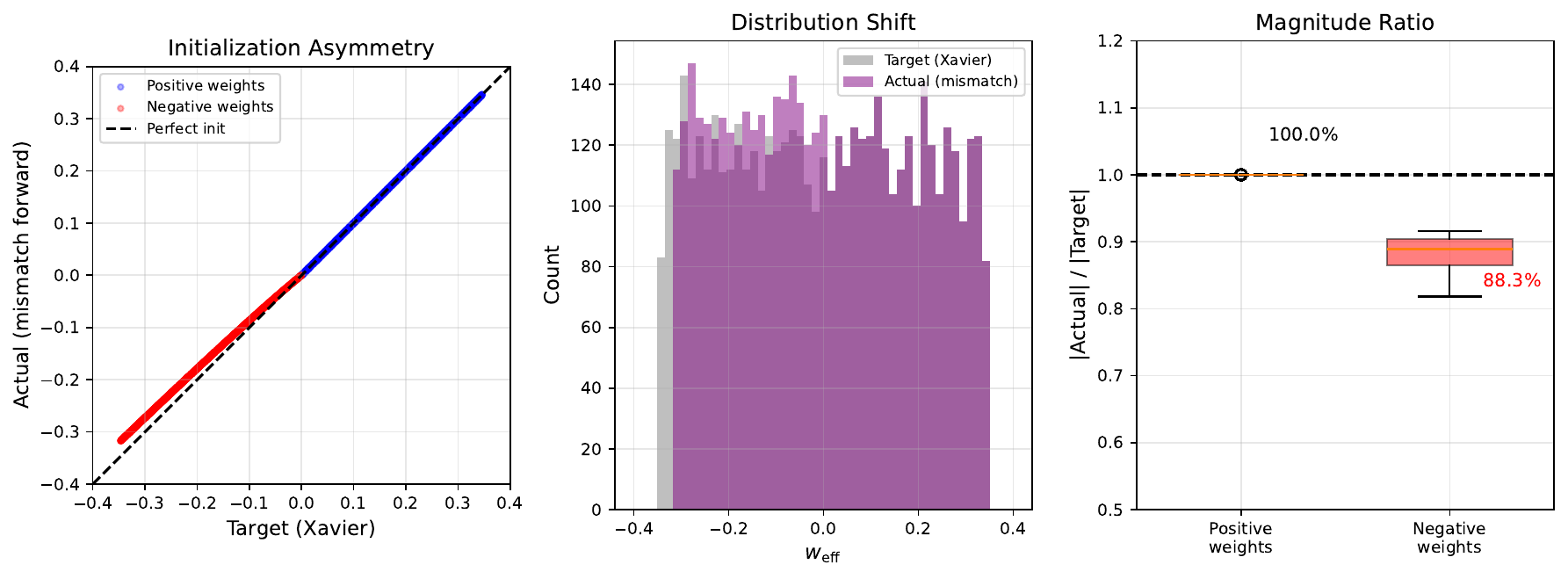}
\caption{\textbf{Detailed analysis of mismatched initialization.} \emph{Left:} Scatter plot of target (Xavier) vs.\ actual effective weights after congruent inversion + mismatch forward. Positive weights (blue) lie on the identity while negative weights (red) are systematically under-initialized. \emph{Middle:} Distribution shift between target and actual. \emph{Right:} Magnitude ratio by sign: positive weights achieve $\sim$100\% of target, negative weights only $\sim$75--80\%.}
\label{fig:mismatched_init_detail}
\end{figure}

\subsection{Weight Distribution Analysis}

\begin{figure}[h]
\centering
\includegraphics[width=0.95\textwidth]{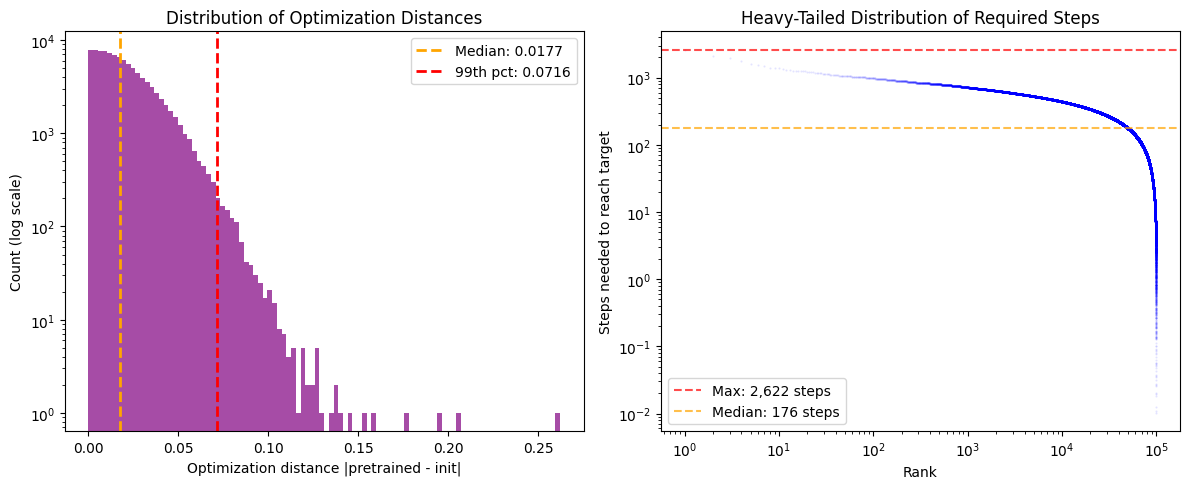}
\caption{\textbf{Optimization distance analysis.} \emph{Left:} Distribution of optimization distances $\abs{w_{\mathrm{pretrained}} - w_{\mathrm{init}}}$ showing a heavy tail. \emph{Right:} Rank plot (log-log) of theoretical minimum steps needed to reach target values, demonstrating that outlier weights require 10--20$\times$ more steps than average.}
\label{fig:opt_distance}
\end{figure}

\begin{figure}[h]
\centering
\includegraphics[width=0.95\textwidth]{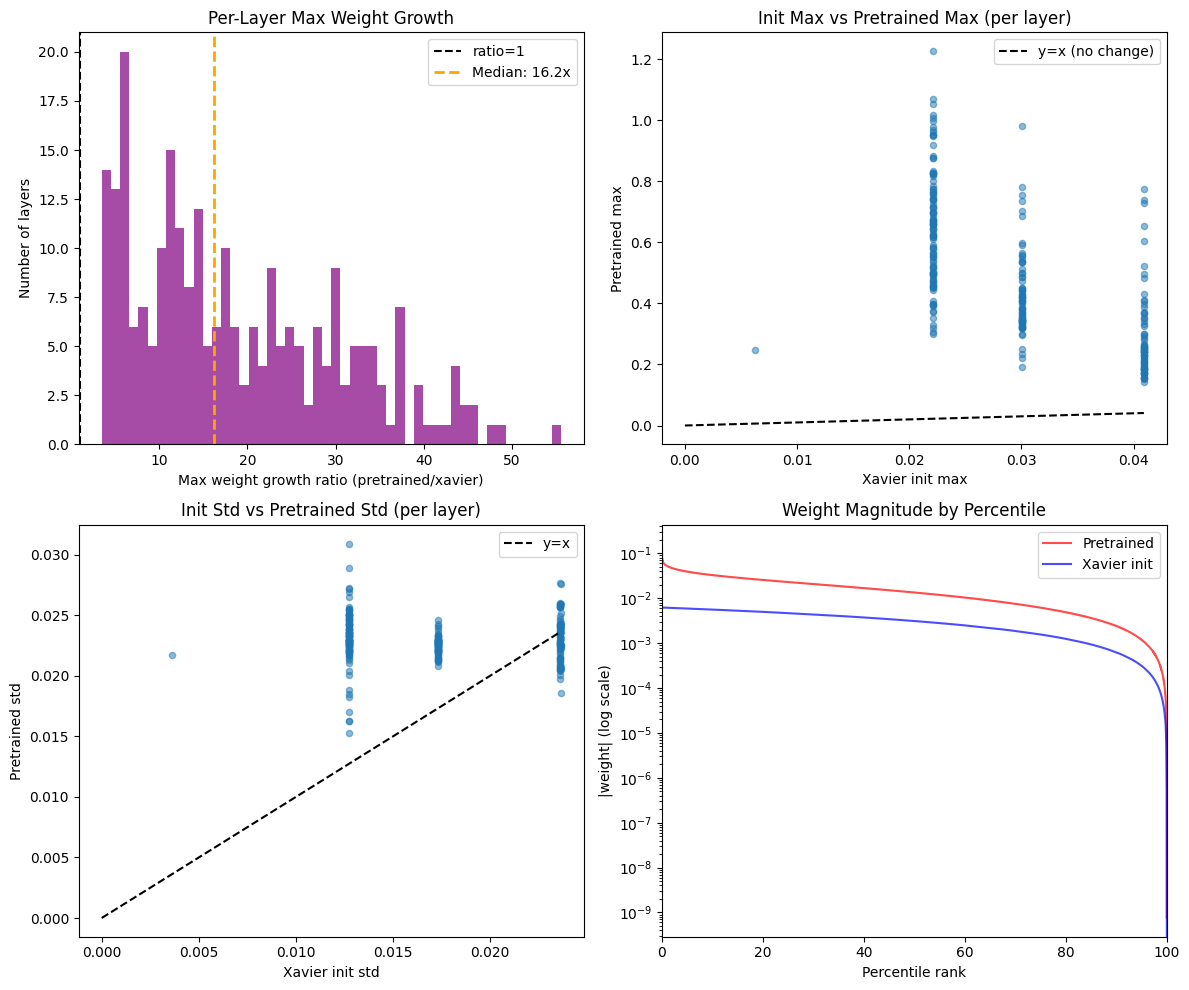}
\caption{\textbf{Heavy-tail characterization of pretrained weights.} \emph{Top left:} Per-layer max weight growth ratio. \emph{Top right:} Correlation between init max and pretrained max per layer. \emph{Bottom left:} Std comparison showing modest overall growth. \emph{Bottom right:} Percentile plot showing the heavy tail. Pretrained weights (red) diverge sharply from Xavier (blue) in the upper percentiles.}
\label{fig:heavy_tail}
\end{figure}

\end{document}